\documentclass{article}

\PassOptionsToPackage{sort&compress}{natbib}

\usepackage[preprint]{neurips_2026}

\usepackage[utf8]{inputenc} 
\usepackage[T1]{fontenc}    
\usepackage[colorlinks]{hyperref}       
\usepackage{url}            
\usepackage{booktabs}       
\usepackage{amsfonts}       
\usepackage{amsmath}

\usepackage{dsfont}
\usepackage{nicefrac}       
\usepackage{etoc}
\makeatletter
\let\origaddcontentsline\addcontentsline
\makeatother
\usepackage{microtype}      
\usepackage{xcolor}         
\usepackage{amssymb}
\usepackage{amsthm}
\usepackage{subcaption}

\usepackage{aliascnt}
\usepackage{cleveref}
\newtheorem{theorem}{Theorem}[section]

\newaliascnt{corollary}{theorem}
\newtheorem{corollary}[corollary]{Corollary}
\aliascntresetthe{corollary}

\newaliascnt{lemma}{theorem}

\aliascntresetthe{lemma}

\newaliascnt{prop}{theorem}
\newtheorem{prop}[prop]{Proposition}
\aliascntresetthe{prop}

\crefname{figure}{Fig.}{Figs.}
\crefname{section}{Sec.}{Secs.}
\crefname{theorem}{Theorem}{Theorems}
\crefname{equation}{Eq.}{Eqs.}
\crefname{corollary}{Corollary}{Corollaries}
\crefname{lemma}{Lemma}{Lemmas}
\crefname{prop}{Proposition}{Propositions}
\crefname{appendix}{App.}{Apps.}
\crefname{table}{Table}{Tables}
\crefname{algorithm}{Algorithm}{Algorithms}

\hypersetup{
  colorlinks=true,
  citecolor=blue,
  linkcolor=red,
  urlcolor=magenta}

\setcitestyle{numbers,open={[},close={]}} 

\usepackage{import}

\usepackage{graphicx}

\DeclareGraphicsExtensions{.pdf,.png}

\usepackage{algorithm}
\usepackage{algpseudocode}

\usepackage{subcaption}
\usepackage{dsfont}
\usepackage{placeins}

\title{Intrinsic Information-Theoretic Analysis of ReLU Nets}

\author{%
  \textbf{Johan Mylius-Kroken}$^{1}$ \quad
  \textbf{Elisabeth Wetzer}$^{1}$ \quad
  \textbf{Ali Ramezani-Kebrya}$^{2}$ \\
  \textbf{Robert Jenssen}$^{1,3,4}$ \quad
  \textbf{Kristoffer Wickstrøm}$^{1}$ \\
  $^{1}$UiT The Arctic University of Norway,
  $^{2}$University of Oslo\\
  $^{3}$Norwegian Computing Center,
  $^{4}$University of Copenhagen\\
  \texttt{johan.m.kroken@uit.no}
}

\makeatletter
\newcommand{\res}[1]{\ifcsname res@#1\endcsname\csname res@#1\endcsname\else\PackageWarning{numbers}{Unknown result key `#1'}\textbf{??\detokenize{#1}}\fi}
\newcommand{\defres}[2]{\expandafter\def\csname res@#1\endcsname{#2}}
\makeatother
\defres{fig2.r.binning8}{0.911}
\defres{fig2.r.kmeansKY}{0.922}
\defres{fig2.r.ksg3}{0.892}
\defres{fig2.n}{75}
\defres{hy.composite.heldout}{2.639}
\defres{hy.mnist.heldout}{3.319}
\defres{hy.wbc.heldout}{0.949}
\defres{hy.composite.insample}{2.639}
\defres{hy.wbc.insample}{0.953}
\defres{fig2.max_bits}{2.835}
\defres{fig2.composite.n}{30}
\defres{fig2.composite.ksg_within_0.02_of_H_Y}{23}
\defres{fig2.composite.ksg_median}{2.631}
\defres{fig2.composite.routing_min}{1.777}
\defres{fig2.composite.routing_max}{2.629}
\defres{mm.n_cells}{138}
\defres{mm.max_correction}{6.488}
\defres{mm.min_value}{-3.169}
\defres{mm.rho<0.05.n_cells}{41}
\defres{mm.rho<0.05.correction}{0.114}
\defres{mm.rho<0.05.plug_in_std}{0.148}
\defres{mm.rho<0.05.mm_std}{0.131}
\defres{mm.rho>0.3.n_cells}{56}
\defres{mm.rho>0.3.correction}{3.739}
\defres{mm.rho>0.3.plug_in_std}{0.025}
\defres{mm.rho>0.3.mm_std}{0.092}
\defres{beyond_mm.rho>0.9.n_cells}{19}
\defres{beyond_mm.rho>0.9.gap_to_H_Y.plug_in}{0.004}
\defres{beyond_mm.rho>0.9.correction.plug_in}{0.000}
\defres{beyond_mm.rho>0.9.gap_to_H_Y.miller_madow}{5.407}
\defres{beyond_mm.rho>0.9.correction.miller_madow}{5.403}
\defres{beyond_mm.rho>0.9.gap_to_H_Y.grassberger}{0.010}
\defres{beyond_mm.rho>0.9.correction.grassberger}{0.006}
\defres{beyond_mm.rho>0.9.gap_to_H_Y.chao_shen}{0.058}
\defres{beyond_mm.rho>0.9.correction.chao_shen}{0.054}
\defres{beyond_mm.rho>0.9.gap_to_H_Y.chao_wang_jost}{0.069}
\defres{beyond_mm.rho>0.9.correction.chao_wang_jost}{0.065}
\defres{beyond_mm.rho>0.9.gap_to_H_Y.ansb}{-9.110}
\defres{beyond_mm.rho>0.9.correction.ansb}{-9.114}
\defres{beyond_mm.ansb_nan_cells}{0}
\defres{beyond_mm.min_N_over_K}{1.001}
\defres{heldout.composite.N_heldout}{2{,}000}
\defres{heldout.composite.N_insample}{10{,}000}
\defres{heldout.composite.max_abs_delta_plug_in.deepest}{0.020}
\defres{heldout.composite.max_abs_delta_plug_in.all_layers}{0.023}
\defres{heldout.composite.max_abs_delta_mm.deepest}{1.059}
\defres{heldout.wbc.N_heldout}{114}
\defres{heldout.wbc.N_insample}{569}
\defres{heldout.wbc.max_abs_delta_plug_in.deepest}{0.031}
\defres{heldout.wbc.max_abs_delta_plug_in.all_layers}{0.075}
\defres{heldout.wbc.max_abs_delta_mm.deepest}{0.152}
\defres{wbc_rho.[25, 25, 25].insample}{0.741}
\defres{wbc_rho.[25, 25, 25].heldout}{0.900}
\defres{wbc_rho.[25, 25, 25, 25, 25].insample}{0.893}
\defres{wbc_rho.[25, 25, 25, 25, 25].heldout}{0.961}
\defres{sensitivity.composite+wbc.n}{60}
\defres{sensitivity.composite+wbc.r_min}{0.655}
\defres{sensitivity.composite+wbc.r_max}{0.976}
\defres{sensitivity.composite+wbc.r.binning2}{0.946}
\defres{sensitivity.composite+wbc.r.binning4}{0.973}
\defres{sensitivity.composite+wbc.r.binning8}{0.945}
\defres{sensitivity.composite+wbc.r.binning16}{0.852}
\defres{sensitivity.composite+wbc.r.binning30}{0.655}
\defres{sensitivity.composite+wbc.r.kmeansKY}{0.970}
\defres{sensitivity.composite+wbc.r.kmeans2KY}{0.974}
\defres{sensitivity.composite+wbc.r.kmeans4KY}{0.974}
\defres{sensitivity.composite+wbc.r.kmeans16}{0.976}
\defres{sensitivity.composite+wbc.r.kmeans64}{0.971}
\defres{sensitivity.composite+wbc.r.kmeans256}{0.964}
\defres{sensitivity.composite+wbc.r.ksg3}{0.944}
\defres{sensitivity.composite+wbc.r.ksg5}{0.948}
\defres{sensitivity.composite+wbc.r.ksg10}{0.953}
\defres{sensitivity.composite+wbc.negative_r}{none}
\defres{sensitivity.composite+wbc.best_binning}{binning4}
\defres{sensitivity.mnist.n}{15}
\defres{sensitivity.mnist.r_min}{-0.888}
\defres{sensitivity.mnist.r_max}{0.875}
\defres{sensitivity.mnist.r.binning2}{0.875}
\defres{sensitivity.mnist.r.binning4}{0.728}
\defres{sensitivity.mnist.r.binning8}{0.690}
\defres{sensitivity.mnist.r.binning16}{-0.570}
\defres{sensitivity.mnist.r.binning30}{-0.888}
\defres{sensitivity.mnist.r.kmeansKY}{0.720}
\defres{sensitivity.mnist.r.kmeans2KY}{0.866}
\defres{sensitivity.mnist.r.kmeans4KY}{0.868}
\defres{sensitivity.mnist.r.kmeans16}{0.834}
\defres{sensitivity.mnist.r.kmeans64}{0.861}
\defres{sensitivity.mnist.r.kmeans256}{0.829}
\defres{sensitivity.mnist.r.ksg3}{0.759}
\defres{sensitivity.mnist.r.ksg5}{0.762}
\defres{sensitivity.mnist.r.ksg10}{0.772}
\defres{sensitivity.mnist.negative_r}{binning16 (r=-0.570), binning30 (r=-0.888)}
\defres{sensitivity.mnist.best_binning}{binning2}
\defres{occupancy.composite.configs}{6}
\defres{occupancy.composite.rho>0.3}{2}
\defres{occupancy.composite.rho_max}{0.411}
\defres{occupancy.mnist.configs}{30}
\defres{occupancy.mnist.rho>0.3}{18}
\defres{occupancy.mnist.rho_max}{1.000}
\defres{occupancy.wbc.configs}{6}
\defres{occupancy.wbc.rho>0.3}{4}
\defres{occupancy.wbc.rho_max}{0.961}
\defres{permutation.composite.[25, 25, 25].clean.raw}{2.535}
\defres{permutation.composite.[25, 25, 25].clean.quotient}{2.362}
\defres{permutation.composite.[25, 25, 25].clean.test_accuracy}{0.995}
\defres{permutation.composite.[25, 25, 25].clean.test_accuracy.pct}{99.5}
\defres{permutation.composite.[25, 25, 25].clean.H_Y}{2.639}
\defres{permutation.composite.[5, 5, 5].clean.raw}{1.949}
\defres{permutation.composite.[5, 5, 5].clean.quotient}{1.828}
\defres{permutation.composite.[5, 5, 5].clean.test_accuracy}{0.979}
\defres{permutation.composite.[5, 5, 5].clean.test_accuracy.pct}{97.9}
\defres{permutation.composite.[5, 5, 5].clean.H_Y}{2.639}
\defres{permutation.wbc.[25, 25, 25].clean.raw}{0.949}
\defres{permutation.wbc.[25, 25, 25].clean.quotient}{0.885}
\defres{permutation.wbc.[25, 25, 25].clean.test_accuracy}{0.972}
\defres{permutation.wbc.[25, 25, 25].clean.test_accuracy.pct}{97.2}
\defres{permutation.wbc.[25, 25, 25].clean.H_Y}{0.949}
\defres{permutation.wbc.[5, 5, 5].clean.raw}{0.767}
\defres{permutation.wbc.[5, 5, 5].clean.quotient}{0.700}
\defres{permutation.wbc.[5, 5, 5].clean.test_accuracy}{0.974}
\defres{permutation.wbc.[5, 5, 5].clean.test_accuracy.pct}{97.4}
\defres{permutation.wbc.[5, 5, 5].clean.H_Y}{0.949}
\defres{permutation.composite.[25, 25, 25].permuted.raw}{2.433}
\defres{permutation.composite.[25, 25, 25].permuted.quotient}{2.329}
\defres{permutation.composite.[25, 25, 25].permuted.test_accuracy}{0.159}
\defres{permutation.composite.[25, 25, 25].permuted.test_accuracy.pct}{15.9}
\defres{permutation.composite.[25, 25, 25].permuted.H_Y}{2.639}
\defres{permutation.composite.[5, 5, 5].permuted.raw}{1.411}
\defres{permutation.composite.[5, 5, 5].permuted.quotient}{1.096}
\defres{permutation.composite.[5, 5, 5].permuted.test_accuracy}{0.196}
\defres{permutation.composite.[5, 5, 5].permuted.test_accuracy.pct}{19.6}
\defres{permutation.composite.[5, 5, 5].permuted.H_Y}{2.639}
\defres{permutation.wbc.[25, 25, 25].permuted.raw}{0.946}
\defres{permutation.wbc.[25, 25, 25].permuted.quotient}{0.932}
\defres{permutation.wbc.[25, 25, 25].permuted.test_accuracy}{0.633}
\defres{permutation.wbc.[25, 25, 25].permuted.test_accuracy.pct}{63.3}
\defres{permutation.wbc.[25, 25, 25].permuted.H_Y}{0.949}
\defres{permutation.wbc.[5, 5, 5].permuted.raw}{0.469}
\defres{permutation.wbc.[5, 5, 5].permuted.quotient}{0.295}
\defres{permutation.wbc.[5, 5, 5].permuted.test_accuracy}{0.625}
\defres{permutation.wbc.[5, 5, 5].permuted.test_accuracy.pct}{62.5}
\defres{permutation.wbc.[5, 5, 5].permuted.H_Y}{0.949}
\defres{permutation.composite.[25, 25, 25].permuted.train_label_accuracy}{0.225}
\defres{permutation.composite.[25, 25, 25].permuted.train_label_accuracy.pct}{22.5}
\defres{permutation.composite.[5, 5, 5].permuted.train_label_accuracy}{0.208}
\defres{permutation.composite.[5, 5, 5].permuted.train_label_accuracy.pct}{20.8}
\defres{permutation.wbc.[25, 25, 25].permuted.train_label_accuracy}{0.656}
\defres{permutation.wbc.[25, 25, 25].permuted.train_label_accuracy.pct}{65.6}
\defres{permutation.wbc.[5, 5, 5].permuted.train_label_accuracy}{0.628}
\defres{permutation.wbc.[5, 5, 5].permuted.train_label_accuracy.pct}{62.8}
\defres{ordering.eps0.3.all_layers.n_cells}{690}
\defres{ordering.eps0.3.all_layers.regions_min}{6}
\defres{ordering.eps0.3.all_layers.regions_max}{9{,}997}
\defres{ordering.eps0.3.all_layers.spread_median}{2}
\defres{ordering.eps0.3.all_layers.spread_max}{84}
\defres{ordering.eps0.3.all_layers.regions_at_spread_max}{6{,}665}
\defres{ordering.eps0.3.all_layers.relative_spread_max}{0.094}
\defres{ordering.eps0.3.all_layers.relative_spread_max.pct}{9.4}
\defres{ordering.eps0.3.all_layers.miller_madow_func_spread_max}{0.148}
\defres{ordering.eps0.3.all_layers.plug_in_func_spread_max}{0.146}
\defres{ordering.eps0.3.deepest.n_cells}{210}
\defres{ordering.eps0.3.deepest.regions_min}{14}
\defres{ordering.eps0.3.deepest.regions_max}{9{,}997}
\defres{ordering.eps0.3.deepest.spread_median}{11}
\defres{ordering.eps0.3.deepest.spread_max}{60}
\defres{ordering.eps0.3.deepest.regions_at_spread_max}{5{,}014}
\defres{ordering.eps0.3.deepest.relative_spread_max}{0.065}
\defres{ordering.eps0.3.deepest.relative_spread_max.pct}{6.5}
\defres{ordering.eps0.3.deepest.miller_madow_func_spread_max}{0.100}
\defres{ordering.eps0.3.deepest.plug_in_func_spread_max}{0.104}
\defres{ordering.n_random_orders}{15}
\defres{ordering.all_eps.miller_madow_func_spread_max}{0.474}
\defres{ordering.all_eps.relative_spread_max}{0.188}
\defres{ordering.all_eps.relative_spread_max.pct}{18.8}
\defres{capacity.H_Y}{3.319}
\defres{capacity.w7.d<=5.raw_max}{2.134}
\defres{capacity.w7.d<=5.gap_min}{1.185}
\defres{capacity.w15.d<=5.raw_max}{2.974}
\defres{capacity.w15.d<=5.gap_min}{0.345}
\defres{capacity.w25.d<=5.raw_max}{3.264}
\defres{capacity.w25.d<=5.gap_min}{0.056}
\defres{capacity.w50.d<=5.raw_max}{3.317}
\defres{capacity.w50.d<=5.gap_min}{0.002}
\defres{capacity.eps0.1.below_raw_max}{0.175}
\defres{capacity.eps0.1.rho_func_min}{0.717}
\defres{capacity.eps0.2.below_raw_max}{0.384}
\defres{capacity.eps0.2.rho_func_min}{0.443}
\defres{capacity.eps0.3.d>=10.w7.gap_min}{1.103}
\defres{capacity.eps0.3.d>=10.w7.gap_max}{1.121}
\defres{capacity.eps0.3.d>=10.w15.gap_min}{0.088}
\defres{capacity.eps0.3.d>=10.w15.gap_max}{0.160}
\defres{capacity.eps0.3.d>=10.w25.gap_min}{0.002}
\defres{capacity.eps0.3.d>=10.w25.gap_max}{0.010}
\defres{capacity.eps0.3.d>=10.w50.gap_min}{0.000}
\defres{capacity.eps0.3.d>=10.w50.gap_max}{0.000}
\defres{capacity.eps0.5.d>=10.w7.gap_min}{1.329}
\defres{capacity.eps0.5.d>=10.w7.gap_max}{1.397}
\defres{capacity.eps0.5.d>=10.w15.gap_min}{0.266}
\defres{capacity.eps0.5.d>=10.w15.gap_max}{0.425}
\defres{capacity.eps0.5.d>=10.w25.gap_min}{0.027}
\defres{capacity.eps0.5.d>=10.w25.gap_max}{0.074}
\defres{capacity.eps0.5.d>=10.w50.gap_min}{0.000}
\defres{capacity.eps0.5.d>=10.w50.gap_max}{0.004}
\defres{capacity.eps2.d2.func_over_raw_min}{0.732}
\defres{capacity.eps2.d2.func_over_raw_max}{0.828}
\defres{capacity.window.max_abs_diff_eps0.3_eps0.5}{0.304}
\defres{rho_func_layerwise.n_networks}{40}
\defres{rho_func_layerwise.n_curves}{560}
\defres{rho_func_layerwise.n_curves_rising}{232}
\defres{rho_func_layerwise.max_rise}{0.216}
\defres{rho_func_layerwise.layer1_all_one}{True}
\defres{ksg_ties.n}{5}
\defres{ksg_ties.datasets}{wbc}
\defres{ksg_ties.N}{114}
\defres{ksg_ties.tied_frac_min}{0.228}
\defres{ksg_ties.tied_frac_min.pct}{22.8}
\defres{ksg_ties.tied_frac_max}{0.781}
\defres{ksg_ties.tied_frac_max.pct}{78.1}
\defres{ksg_ties.n_pure}{4}
\defres{ksg_ties.n_dead_upstream}{5}
\defres{ksg_ties.n_ksg_raw_negative}{4}
\defres{ksg_ties.ksg_max}{0.148}
\defres{ksg_ties.routing_min}{0.392}
\defres{ksg_ties.routing_max}{0.911}
\defres{ksg_ties.binning_min}{0.217}
\defres{ksg_ties.binning_max}{0.795}
\defres{ksg_ties.untied.composite.tied_frac_max}{0.077}
\defres{ksg_ties.untied.composite.tied_frac_max.pct}{7.7}
\defres{ksg_ties.untied.mnist.tied_frac_max}{0.053}
\defres{ksg_ties.untied.mnist.tied_frac_max.pct}{5.3}
\defres{ksg_ties.untied.wbc.tied_frac_max}{0.158}
\defres{ksg_ties.untied.wbc.tied_frac_max.pct}{15.8}
\defres{ksg_ties.untied.wbc.arch}{[5, 5, 5]}
\defres{ksg_ties.untied.wbc.seed}{105}
\defres{ksg_ties.untied.wbc.ksg3}{0.436}
\defres{ksg_ties.fig2.r_ksg3.with}{0.892}
\defres{ksg_ties.fig2.r_ksg3.without}{0.893}
\defres{ksg_ties.fig2.n_without}{70}
\defres{ksg_ties.fig3.wbc.n_tied}{4}
\defres{ksg_ties.fig3.wbc.onset_layer_min}{3}
\defres{ksg_ties.fig3.wbc.onset_layer_max}{3}
\defres{ksg_ties.fig3.wbc.ksg3.with.l1}{0.824}
\defres{ksg_ties.fig3.wbc.ksg3.with.l2}{0.837}
\defres{ksg_ties.fig3.wbc.ksg3.with.l3}{0.643}
\defres{ksg_ties.fig3.wbc.ksg3.with.l4}{0.619}
\defres{ksg_ties.fig3.wbc.ksg3.with.l5}{0.623}
\defres{ksg_ties.fig3.wbc.ksg3.with.min}{0.619}
\defres{ksg_ties.fig3.wbc.ksg3.with.max}{0.837}
\defres{ksg_ties.fig3.wbc.ksg3.without.l1}{0.823}
\defres{ksg_ties.fig3.wbc.ksg3.without.l2}{0.840}
\defres{ksg_ties.fig3.wbc.ksg3.without.l3}{0.835}
\defres{ksg_ties.fig3.wbc.ksg3.without.l4}{0.833}
\defres{ksg_ties.fig3.wbc.ksg3.without.l5}{0.835}
\defres{ksg_ties.fig3.wbc.ksg3.without.min}{0.823}
\defres{ksg_ties.fig3.wbc.ksg3.without.max}{0.840}
\defres{ksg_ties.fig3.wbc.routing.with.l1}{0.598}
\defres{ksg_ties.fig3.wbc.routing.with.l2}{0.775}
\defres{ksg_ties.fig3.wbc.routing.with.l3}{0.825}
\defres{ksg_ties.fig3.wbc.routing.with.l4}{0.844}
\defres{ksg_ties.fig3.wbc.routing.with.l5}{0.854}
\defres{ksg_ties.fig3.wbc.routing.with.min}{0.598}
\defres{ksg_ties.fig3.wbc.routing.with.max}{0.854}
\defres{ksg_ties.fig3.wbc.routing.without.l1}{0.675}
\defres{ksg_ties.fig3.wbc.routing.without.l2}{0.815}
\defres{ksg_ties.fig3.wbc.routing.without.l3}{0.873}
\defres{ksg_ties.fig3.wbc.routing.without.l4}{0.897}
\defres{ksg_ties.fig3.wbc.routing.without.l5}{0.905}
\defres{ksg_ties.fig3.wbc.routing.without.min}{0.675}
\defres{ksg_ties.fig3.wbc.routing.without.max}{0.905}

\begin{document}

\maketitle

\begin{abstract}
	Mutual information (MI) is fundamental to representation-learning analysis in deep networks, but in high dimensions MI estimators depend on hyperparameters whose bias often dominates the signal. We show that continuous piecewise-linear (CPWL) networks, including ReLU networks, admit an intrinsic information-theoretic analysis that avoids this pitfall: their inherent partition of the input space into convex linear regions assigns to every input a discrete region label $\Pi$ alongside the continuous representation $T$ and label $Y$.  We introduce a decomposition of the mutual information $I(Y;T)$ into interpretable geometric terms. We show that the dominant term, the \emph{routing information} $I(Y; \Pi)$, admits an estimable plug-in estimator, that in the well-sampled regime, approximates a population lower bound on the joint information $I(Y; \Pi, T)$. When the partition saturates, a functional-equivalence quotient collapses regions implementing the same linear operator and restores informativeness; this extension has a single bounded, dimensionless parameter, which is reported through the quotient's compression ratio. The routing estimator agrees strongly with three established MI baselines applied to the same networks and data. The functional quotient recovers informative estimates in the regime bordering the well-sampled and under-sampled regime. We provide a framework for analysing the internal information geometry of CPWL networks, grounded in their inherent piecewise-linear structure rather than in external discretisation. Project page:\newline \url{https://johanmkr.github.io/Intrinsic-IT-analysis-of-ReLU-nets/}
\end{abstract}

\section{Introduction}
\label{sec:introduction}
Information-theoretic accounts of neural networks span training dynamics \citep{shwartz-zivOpeningBlackBox2017,adilovaInformationPlaneAnalysis2022}, loss design \citep{belghaziMutualInformationNeural2018,jeonInformationTheoreticDiscreteDiffusion2025}, and representation analysis \citep{goldfeldEstimatingInformationFlow2019,felUnderstandingVisualFeature2024b}, but their empirical validation is brittle: classical non-parametric estimators (binning, kernel density, $k$-nearest neighbours \citep{moonEstimationMutualInformation1995,kraskovEstimatingMutualInformation2004}) and variational neural estimators \citep{belghaziMutualInformationNeural2018,huInfoNetNeuralEstimation2024a} all depend on user-chosen hyperparameters --- bin widths, bandwidths, neighbourhood radii, or auxiliary architectures --- whose biases dominate the signal in deep networks \citep{paninskiEstimationEntropyMutual2003a,goldfeldEstimatingInformationFlow2019,geigerInformationPlaneAnalyses2022}. Can we compute information-theoretic quantities \emph{intrinsic} to the network, without such choices?

Continuous Piecewise Linear (CPWL) networks \citep{haninComplexityLinearRegions2019,goujonNumberRegionsPiecewise2023} --- including ReLU networks \citep{glorotDeepSparseRectifier2011} --- partition the input space $\mathcal{X}$ into convex \emph{linear regions} on which the network acts as an affine map (\cref{fig:pedagogical_figure}). Each region $\omega$ is uniquely identified by the binary activation pattern $\pi_\omega$ obtained by thresholding pre-activations at zero, so every input carries a discrete region label $\Pi$ alongside its continuous representation $T = \phi_\Pi(X)$, and information-theoretic quantities involving $\Pi$ are well-defined on a finite alphabet without any discretisation choice. We use this to develop a framework for the geometry-based information of CPWL networks. The main contributions are:

\begin{figure}[h]
	\centering
	\includegraphics[width=0.95\textwidth]{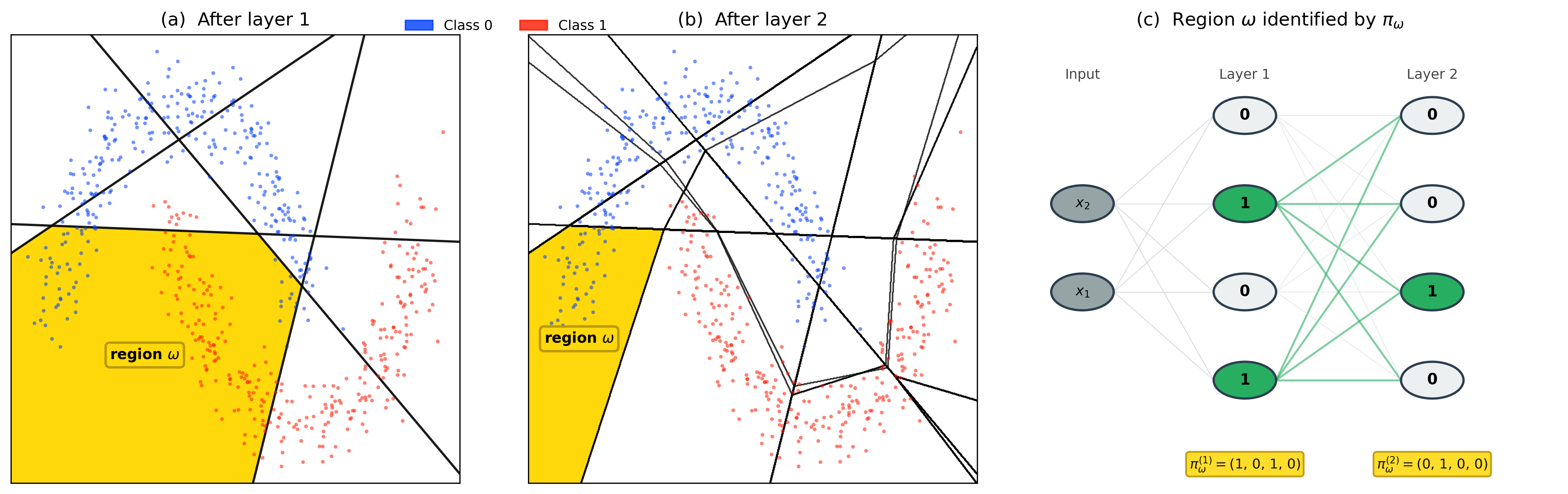}
	\caption{From geometry to combinatorics in a ReLU network.
		\textbf{(a)}~The first layer partitions the input space
		$\mathcal{X} \subset \mathbb{R}^2$ into convex linear regions;
		black lines are the projections of the layer's neuron hyperplanes
		onto $\mathcal{X}$. A single region $\omega$ is highlighted in
		yellow. \textbf{(b)}~Composing the second layer refines the
		partition: each new hyperplane subdivides regions of~(a), so the
		partition through layer~$2$ is the \emph{common refinement} of
		the per-layer partitions and $\omega$ itself is split. The
		network is affine on every region of this refinement.
		\textbf{(c)}~The same region $\omega$ is uniquely identified by
		its activation pattern
		$\pi_\omega = (\pi^{(1)}_\omega, \pi^{(2)}_\omega)
			= ((1,0,1,0), (0,1,0,0))$ --- the layer-wise binarisation of
		the pre-activations (green: active, white: inactive). The
		bijection between regions and activation patterns is the bridge
		between the geometric and combinatorial views of CPWL networks
		that our framework exploits for information-theoretic analysis.}
	\label{fig:pedagogical_figure}
\end{figure}

\begin{itemize}
	\item \textbf{Information decomposition} \cref{thm:cpwl_decomp} decomposes $I(Y;T)$ into geometrically interpretable parts, where one is a discrete and estimable measure of \textit{routing information}. Other terms are described in \cref{prop:folding_loss,cor:upper_bound,cor:balance}.

	\item \textbf{Routing estimator} (\cref{subsec:estimator}). The plug-in estimator is computed from the partition and, in the well-sampled regime, approximates $I(Y;\Pi)$. When it saturates, the functional-equivalence quotient (\cref{prop:func_quotient}) introduces a coarsening of $\Pi$ that merges regions implementing nearly identical linear operators under a Frobenius norm criterion. This extends the estimator into the fine-resolution regime where the number of regions approaches the number of samples.
\end{itemize}

We complement the theory with an empirical study (\cref{sec:results}) on three structurally distinct datasets. Its calibration part is an internal-consistency check of the estimator: does the plug-in estimate on the network's own routing alphabet agree with three established MI baselines applied to the same networks and data? It does not test whether $I(Y;\Pi)$ approximates $I(Y;T)$; the two can differ in either direction (\cref{cor:balance}, \cref{sec:calibration-layerwise}). A saturation-correction experiment then shows that the functional-equivalence quotient recovers meaningful routing estimates at growing input dimensions.

\paragraph{Implications.}
Replacing aggregate $I(Y;T)$ with its decomposition turns hyperparameter-conditioned IB observations on CPWL networks --- whether information rises or falls with depth, whether folding destroys class content, whether MI saturates before generalisation --- into measurable claims about specific terms. This is a first step towards an information-theoretic understanding of neural networks in which the base estimate originates from the network geometry.


\section{Related Work}
\label{sec:related_work}

\paragraph{Geometry of CPWL networks.}
The combinatorial study of linear regions in ReLU networks was established by \citet{NIPS2014_fa6f2a46} and \citet{pascanuNumberResponseRegions2013}, with refined bounds and empirical counts in \citet{serraBoundingCountingLinear2018,haninComplexityLinearRegions2019} -- the latter showing that practical region counts are far below the theoretical maximum -- and generalised to arbitrary CPWL activations via convex-partition arrangements by \citet{goujonNumberRegionsPiecewise2023}, the framework we adopt in \cref{subsec:prelim}. Algorithmic extraction has been pursued through 2D-specialised exact visualisation \citep{humayunSplineCamExactVisualization2024}, edge-subdivision \citep{berzinsPolyhedralComplexExtraction2023}, and polyhedral marching that prunes unreachable regions \citep{vincentReachablePolyhedralMarching2021,liuIntegratingGeometriesReLU2023a,sattelbergLocallyLinearAttributes2023}. \textbf{MI estimation in deep networks.}
The information-bottleneck framework \citep{tishbyInformationBottleneckMethod2000a,parkUnsupervisedRepresentationLearning,tishbyDeepLearningInformation2015a,shwartz-zivOpeningBlackBox2017} views representation learning as a tradeoff between $I(X;T)$ and $I(Y;T)$, but \citet{saxeInformationBottleneckTheory2019} and \citet{goldfeldEstimatingInformationFlow2019} showed that $I(Y;T)$ is ill-defined for deterministic networks on continuous inputs, with estimates dominated by binning and noise-injection choices. Alternative formulations include sliced MI \citep{goldfeldSlicedMutualInformation2021,wongsoUsingSlicedMutual2023} and variational lower bounds \citep{belghaziMutualInformationNeural2018,oordRepresentationLearningContrastive2018}. These replace $I(Y;T)$ by a better-behaved surrogate; sliced MI, for instance, averages the MI of random one-dimensional projections and adds a choice of slice distribution and count. We instead exploit the discrete partition $\Pi$ inherent to CPWL networks, on which $I(Y; \Pi)$ is well-defined on finite alphabets and estimable without such choices. It is not a surrogate for $I(Y;T)$ but one term of its exact decomposition (\cref{thm:cpwl_decomp}): it lower-bounds $I(Y;\Pi,T)$, and its gap to $I(Y;T)$ is the difference between the within-region information and a routing-loss term $I(Y; \Pi \mid T)$ (\cref{cor:balance}) that has no analogue in the standard IB framework. Sliced MI on the same networks would be a complementary view of $I(Y;T)$, which we leave to future work. \textbf{Activation-pattern analysis.} It is established that the number of activation patterns encountered in practice is far below the theoretical maximum \cite{haninDeepReLUNetworks2019}, motivating the data-restricted analysis. \citet{lewandowskiExploitingSpaceFolding2026} introduced a space-folding measure based on Hamming-distance non-monotonicity along input-space paths, and \citet{dhayalkarGeometryReLUNetworks2025} formalised the global combinatorial structure as the \emph{ReLU Transition Graph}, and later extended it with region-entropy and KL-based analysis \cite{dhayalkar2025discretefunctionalgeometryrelu}, further put into a graph framework by \citet{gaines2026characterizing}.

\section{Intrinsic information theoretic analysis of CPWL networks}
\label{sec:framework}



The inherent partition of CPWL networks assigns every input both a continuous representation $T$ and a discrete region label $\Pi$, and the label takes finitely many values regardless of what the per-region affine maps do. This section uses that structure: we set up notation (\cref{subsec:prelim}), propose the exact information decomposition (\cref{subsec:decomp}) with its structural consequences (\cref{subsec:inequalities}), and turn to estimation: a plug-in routing estimator (\cref{subsec:estimator}) and a functional-equivalence quotient that extends its reach into the fine-resolution regime (\cref{subsec:functional_quotient}).

\subsection{Preliminaries}
\label{subsec:prelim}

A CPWL neural network $\mathcal{N}: \mathcal{X} \to \mathcal{Y}$ is the composite function $\mathbf{y} = \mathcal{N}(\mathbf{x}) = (\sigma_L \circ f_{\theta_L} \circ \dots \circ \sigma_1 \circ f_{\theta_1})(\mathbf{x})$,
mapping input $\mathbf{x}$ to output $\mathbf{y}$. It has affine layers $f_{\theta_l}: \mathbb{R}^{n_{l-1}} \to \mathbb{R}^{n_l}$ parametrized by $\theta_l = \{\mathbf{W}^l \in \mathbb{R}^{n_l \times n_{l-1}}, \mathbf{b}^l \in \mathbb{R}^{n_l}\}$ followed by point-wise ReLU activation $\sigma(z) = \max(0, z)$. We denote $\mathcal{N}^l$ the sub-network truncated after layer $l$, write $\mathbf{z}^l = \mathbf{W}^l \mathbf{t}^{l-1} + \mathbf{b}^l$ for its pre-activation and $\mathbf{t}^l = \sigma_l(\mathbf{z}^l) = \mathcal{N}^l(\mathbf{x})$ for its (post-activation) output. When unambiguous, we drop the superscript $l$.

\paragraph{Regions, activation patterns and partition.} A \emph{region} $\omega \subset \mathcal{X}$ is a subset of the input domain on which $\mathcal{N}^l$ is a single affine function. Each region in layer $l$ is uniquely identified by an 
\emph{activation pattern} $\pi_\omega = (\pi^1_\omega, \dots, \pi^{l}_\omega)$, the concatenation of per-layer binarisations $\pi^l_\omega = \mathbf{1}[\mathbf{z}^l > 0] \in \{0, 1\}^{n_l}$  for layers $1,\ldots,l$, obtained by thresholding the layer-$l$ pre-activations at zero (\cref{fig:pedagogical_figure}, panel c).
We write $\Pi^l$ for the \emph{collection} of regions induced by the network through layer $l$, so that $\bigcup_{\omega \in \Pi^l} \omega = \mathcal{X}$ partitions the input domain. In practice, only a small fraction of $\Pi^l$ contain data \citep{haninComplexityLinearRegions2019}, and all our estimators are evaluated on this subset. Given a dataset $\mathcal{D} \subset \mathcal{X}$ of $N := |\mathcal{D}|$ samples, the \emph{data-supported regions} are $\Omega^l_{\mathcal{D}} = \{\, \omega \in \Pi^l : \omega \cap \mathcal{D} \neq \varnothing \,\}$
with cardinalities $|\Omega^l_{\mathcal{D}}| \leq \min\{|\Pi^l|,|\mathcal{D}|\}$. Within any $\omega \in \Pi^l$ the network behaves as an affine map $\phi^l_\omega:\omega\to\mathbb{R}^{n_l}$ such that $\mathcal{N}^l|_\omega = \phi^l_\omega$, with
\begin{equation}\label{eq:per_region_affine}
	\phi^l_\omega(\mathbf{x}) = \mathbf{A}^l_\omega \mathbf{x} + \mathbf{c}^l_\omega,
	\quad
	\mathbf{A}^l_\omega = \prod_{i=l}^{1} \mathbf{Q}^i_\omega \mathbf{W}^i,
	\quad
	\mathbf{c}^l_\omega = \mathbf{Q}^l_\omega \mathbf{b}^l
	+ \sum_{j=1}^{l-1}
	\left(\prod_{k=l}^{j+1} \mathbf{Q}^k_\omega \mathbf{W}^k\right) \mathbf{Q}^j_\omega \mathbf{b}^j,
\end{equation}
where $\prod_{i=l}^{1}$ denotes the descending product $\mathbf{Q}^l_\omega \mathbf{W}^l \cdots \mathbf{Q}^1_\omega \mathbf{W}^1$ (matching feed-forward composition) and $\mathbf{Q}^i_\omega = \mathrm{diag}(\pi^i_\omega)$ selects the active neurons in layer $i$. Construction of $\Omega^l_{\mathcal{D}}$ from a trained network and a dataset reduces to evaluating $\pi_\omega(\mathbf{x})$ at every $\mathbf{x} \in \mathcal{D}$ and grouping by pattern (\cref{app:linear_regions}).

\subsection{Probabilistic setup for information bottleneck chain}
\label{subsec:prob_setup}

We treat the input $X$ and label $Y$ as random variables with $|\mathcal{Y}| < \infty$. For layer $l$ and a CPWL with given parameters, both $\Pi = \pi(X) \in \Pi^l$ and $T = \phi(X) \in \mathbb{R}^{n_l}$ are deterministic functions of $X$. The standard bottleneck chain $Y \to X \to T$ lifts to the directed acyclic graph (DAG): $Y \longrightarrow X \longrightarrow \Pi, \quad X \longrightarrow T, \quad \Pi \longrightarrow T$
in which $\Pi$ and $T$ are \emph{parallel} descendants of $X$, not sequential. We use two conditional independencies: $Y \perp (\Pi, T) \mid X$ and $H(\Pi \mid X) = H(T \mid X) = 0$. Neither $X \to \Pi \to T$ nor $X \to T \to \Pi$ holds in general: within a region $T$ is nontrivial in $X$, and distinct regions can share $T$-values under rank-deficient $\mathbf{A}_\omega$. The special case $l = 1$, where $\Pi = \mathbf{1}[T > 0]$, is the only setting in which $T$ determines $\Pi$.

We write $\nu_\omega := p_{T \mid \Pi}(\,\cdot\, \mid \omega)$ for the \emph{within-region latent-space density} -- the density of $T$ inside region $\omega$. By the affine map of~\cref{eq:per_region_affine}, $\nu_\omega$ is supported on the affine image $\phi_\omega(\omega) \subset \mathbb{R}^{n_l}$ (possibly singular when $\mathbf{A}_\omega$ is rank-deficient), and the marginal density of $T$ is the mixture $p_T = \sum_\omega p_\Pi(\omega)\,\nu_\omega$. A formal pushforward derivation and rank discussion are in \cref{app:joint}.

\subsection{Decomposition of $I(Y; T)$}
\label{subsec:decomp}

\begin{theorem}[CPWL Information Decomposition]\label{thm:cpwl_decomp}
	From the DAG we extract the following:
	\begin{equation}\label{eq:main_decomp}
		I(Y; T) = \underbrace{I(Y; \Pi)}_{\text{routing}}
		+ \underbrace{I(Y; T \mid \Pi)}_{\text{within-region}}
		- \underbrace{I(Y; \Pi \mid T)}_{\text{routing loss}}
	\end{equation}
	and
	\begin{equation}\label{eq:joint_decomp}
		I(Y; \Pi, T) = I(Y; \Pi) + I(Y; T \mid \Pi)
		= I(Y; X) - I(Y; X \mid \Pi, T).
	\end{equation}
\end{theorem}

The decomposition follows from chain-ruling $I(Y; \Pi, T)$ in two ways (proof in \cref{app:proof_cpwl_decomp}). Each term has a clean interpretation:
\begin{itemize}
	\item $I(Y; \Pi)$ -- \emph{routing information}: task information in which region the input lands. Finite, well-defined, and the only directly estimable term (\cref{subsec:estimator}).
	\item $I(Y; T \mid \Pi) = \sum_\omega p_\Pi(\omega)\, I_\omega(Y; \mathbf{A}_\omega X)$ -- \emph{within-region information}: how well $\mathrm{row}(\mathbf{A}_\omega)$ aligns with $Y$-relevant directions of $X$.
	\item $I(Y; \Pi \mid T)$ -- \emph{routing loss}: class-relevant region information destroyed by reading $T$ alone; it is the information signature of folding (\cref{prop:folding_loss}).
	\item $I(Y; X \mid \Pi, T)$ -- \emph{representation insufficiency}: task information invisible to $(\Pi, T)$; zero iff $(\Pi, T)$ is sufficient for $Y$.
\end{itemize}
Per-region expressions for all four terms are in \cref{app:per_region}.

\paragraph{Overlapping latent supports and routing loss.}
The routing-loss term has a clean geometric reading: it is positive precisely when distinct regions place mass on the same $t \in \mathbb{R}^{n_l}$ \emph{and} disagree about the class distribution there.

\begin{prop}[Routing loss from latent overlap]\label{prop:folding_loss}
	$I(Y; \Pi \mid T) > 0$ iff there exist distinct regions $\omega, \omega' \in \Pi^l$ with $\mathrm{supp}(\nu_\omega) \cap \mathrm{supp}(\nu_{\omega'}) \neq \varnothing$ of positive $p_T$-mass, \emph{and} whose conditional label distributions differ on the overlap, $p_{Y \mid \Pi = \omega, T = t} \neq p_{Y \mid \Pi = \omega', T = t}$ on a positive-mass subset.
\end{prop}

A simpler sufficient condition, satisfied whenever within-region label uncertainty is independent of $T$, is that two such overlapping regions have differing region-conditional label laws $p_{Y \mid \Pi = \omega} \neq p_{Y \mid \Pi = \omega'}$. Proof and remarks are in \cref{app:proof_folding_loss}.

\subsection{Structural inequalities}
\label{subsec:inequalities}

\begin{corollary}[Upper bound]\label{cor:upper_bound}
	$I(Y; T) \leq I(Y; \Pi, T) \leq I(Y; X)$, with right equality iff $(\Pi, T)$ is sufficient for $Y$, and left equality iff $I(Y; \Pi \mid T) = 0$.
\end{corollary}

\begin{corollary}[Routing vs.\ within-region balance]\label{cor:balance}
	$I(Y; T) - I(Y; \Pi) = I(Y; T \mid \Pi) - I(Y; \Pi \mid T)$. In particular, $I(Y; T) > I(Y; \Pi)$ iff within-region discrimination exceeds routing loss.
\end{corollary}

Unlike the standard information-bottleneck chain, neither of $I(Y; \Pi)$ and $I(Y; T)$ dominates unconditionally; the sign of their difference is a measurable property of the network.

\subsection{The plug-in routing estimator}
\label{subsec:estimator}

Of the four terms in \cref{thm:cpwl_decomp}, only $I(Y; \Pi)$ is a discrete MI on finite alphabets and directly estimable. Given counts $n_{y, \omega}, n_y, n_\omega$ over the data-supported regions $\Omega^l_{\mathcal{D}}$, the plug-in estimator is
\begin{equation}\label{eq:plugin}
	\hat{I}(Y; \Omega_{\mathcal{D}})
	= \sum_{y, \omega} \frac{n_{y, \omega}}{N}
	\log_2 \frac{n_{y, \omega}\, N}{n_y\, n_\omega},
	\qquad N = |\mathcal{D}|.
\end{equation}
In the well-sampled regime $\Pr(\Pi \notin \Omega_{\mathcal{D}}) \ll 1$, $\hat{I}$ approximates $I(Y; \Pi)$, and the chain
\begin{equation}\label{eq:chain}
	\hat{I}(Y; \Omega_{\mathcal{D}})
	\;\approx\; I(Y; \Pi)
	\leq I(Y; \Pi, T)
	\leq I(Y; X)
	\leq H(Y)
\end{equation}
holds up to the truncation gap, bounded by $H(Y)\Pr(\Pi \notin \Omega_{\mathcal{D}}) + h_2(\Pr(\Pi \notin \Omega_{\mathcal{D}}))$ with $h_2$ the binary entropy (\cref{app:proof_routing_est}). $\hat{I}$ is the only quantity that needs to be estimated; every inequality has a clean interpretation through \cref{thm:cpwl_decomp}. The plug-in estimator is biased upwards at finite $N$. We compare it with four standard bias correction, Miller-Madow~\cite{miller1955note}, Grassberger~\cite{grassbergerEntropyEstimatesInsufficient2003}, Chao-Shen~\cite{chaoNonparametricEstimationShannon2003} and Chao-Wang-Jost~\cite{chaoEntropySpeciesAccumulation2013} (\cref{app:bias_corrections}).

\paragraph{Target and occupancy.}
The chain uses ``$\approx$'' rather than ``$\leq$'' because $\hat{I}$, whose empirical frequencies are normalised over the observed regions, targets $I(Y; \Pi \mid \Pi \in \Omega_{\mathcal{D}})$. Conditioning on the observed regions can move MI in either direction, so $\hat{I}$ is not in general a population lower bound for $I(Y; \Pi)$ (\cref{app:proof_routing_est}). To judge which regime an estimate is in, we report the \emph{occupancy}
\begin{equation}\label{eq:rho}
	\rho^l := |\Omega^l_{\mathcal{D}}| \,/\, N \;\in\; (0, 1]
\end{equation}
together with the \emph{singleton fraction}, the share of regions that hold a single sample. Both are read off the count table of \cref{eq:plugin}. At small $\rho$ most regions hold several points and $\hat{I}$ is close to $I(Y; \Pi)$. As $\rho \to 1$ almost every region is a singleton, the partition becomes an injective map from samples to regions, and $\hat{I}$ approaches the empirical label entropy whether or not the network has learned class structure. Estimates in this regime are not reliable, and standard bias corrections do not repair them (\cref{app:bias_corrections}). The occupancy $\rho$ is distinct from the compression ratio $\rho_{\mathrm{func}}$ of \cref{subsec:functional_quotient}.

\paragraph{Computational cost.}
Constructing $\Omega^l_{\mathcal{D}}$ takes one forward pass per sample, $O(N \cdot |\theta_{\leq l}|)$, which is the ordinary inference cost. Hashing the activation patterns costs $O(N \sum_{i \leq l} n_i)$, and filling the count table and evaluating \cref{eq:plugin} costs $O(N)$ amortised plus $O(|\Omega_{\mathcal{D}}| \cdot |\mathcal{Y}|)$, with $|\Omega_{\mathcal{D}}| \leq N$. The estimator therefore adds overhead linear in $N$ to a forward pass that is needed anyway, and is asymptotically cheaper than the pairwise clustering of the functional quotient (\cref{app:func_algorithm}).

\subsection{The fine-resolution regime and the functional quotient}
\label{subsec:functional_quotient}

\cref{eq:chain} is informative when $N \gg |\Omega_{\mathcal{D}}|$. For typical networks this fails: $\pi_\omega \in \{0, 1\}^{\sum_i n_i}$ admits exponentially many patterns and at modest depth $|\Omega_{\mathcal{D}}|$ approaches $N$. In that regime $\hat{I}$ saturates at $H(Y)$ and the bound is vacuous. The issue is structural: the partition $\Omega_{\mathcal{D}}$ becomes too fine to be meaningfully estimated from $\mathcal{D}$. We need a coarsening of $\Pi$ that lower-bounds $I(Y; \Pi)$ and exploits the affine structure of~\cref{eq:per_region_affine}. We propose to group regions that implement the same linear operator.

\paragraph{Functional equivalence.}
Two regions are \emph{functionally equivalent} if they implement the same linear operator: $\omega \sim_{\mathrm{func}} \omega' \;:\iff\; \mathbf{A}^l_\omega = \mathbf{A}^l_{\omega'}$. The bias $\mathbf{c}^l_\omega$ is a region-dependent offset and is excluded. The quotient is denoted $\Omega_{\mathcal{D}, \mathrm{func}} = \Omega^l_{\mathcal{D}} / \sim_{\mathrm{func}}$.

\begin{prop}[Tight lower bound via functional quotient]\label{prop:func_quotient}
	$I(Y; \Omega_{\mathcal{D}, \mathrm{func}}) \leq I(Y; \Pi)$, with equality iff $p_{Y \mid \Pi = \omega}$ depends on $\omega$ only through $\mathbf{A}^l_\omega$.
\end{prop}

Proof and remarks (predicted vs.\ ground-truth label) are in \cref{app:proof_func_quotient}.

\paragraph{Computing the quotient.}
Materializing each $\mathbf{A}^l_\omega$ is expensive. We use the active sub-network matrix $\tilde{\mathbf{A}}^l_\omega = \prod_{i=l}^{1} \mathbf{W}^i[S^\omega_i, S^\omega_{i-1}]$, where $S^\omega_i \subseteq \{1, \dots, n_i\}$ is the set of active neurons in layer $i$, read directly from $\pi^i_\omega$, with $S^\omega_0 := \{1, \dots, n_0\}$, and $\mathbf{W}^i[S^\omega_i, S^\omega_{i-1}]$ is the corresponding submatrix of $\mathbf{W}^i$. $\mathbf{A}^l_\omega$ is obtained from $\tilde{\mathbf{A}}^l_\omega$ by inserting zero rows at inactive output positions. Since $\tilde{\mathbf{A}}^l_\omega$ is a product of $l$ matrices, its Frobenius norm scales with $\prod_i\|\mathbf{W}^i\|$ and varies with depth. We therefore use the \emph{relative} Frobenius criterion at tolerance $\varepsilon \geq 0$,
\begin{equation}\label{eq:func_equiv}
	\omega \sim_\varepsilon \omega'
	\;:\iff\;
	S^\omega_l = S^{\omega'}_l
	\;\;\text{and}\;\;
	\frac{\|\tilde{\mathbf{A}}^l_\omega - \tilde{\mathbf{A}}^l_{\omega'}\|_F}
	{\tfrac{1}{2}\big(\|\tilde{\mathbf{A}}^l_\omega\|_F + \|\tilde{\mathbf{A}}^l_{\omega'}\|_F\big)}
	\leq \varepsilon,
\end{equation}
which is dimensionless, bounded in $[0, 2]$, and scale-invariant across layers and architectures. The endpoints have clean meaning: $\varepsilon = 0$ recovers the raw partition; $\varepsilon \geq 1$ merges most regions regardless of architecture. \Cref{sec:capacity-results} identifies $\varepsilon \in [0.2, 0.5]$ as the useful range. Details are in \cref{app:func_algorithm}.

\paragraph{The functional-quotient estimator.}
~\cref{eq:plugin} on $\Omega_{\mathcal{D}, \mathrm{func}}$ yields $\hat{I}_{\mathrm{func}} := \hat{I}(Y; \Omega_{\mathcal{D}, \mathrm{func}})$ satisfying
\begin{equation}\label{eq:func_chain}
	\hat{I}_{\mathrm{func}}
	\;\approx\; I(Y; \Omega_{\mathcal{D}, \mathrm{func}})
	\leq I(Y; \Pi)
	\leq I(Y; \Pi, T)
	\leq I(Y; X)
	\leq H(Y),
\end{equation}
extending the chain~\cref{eq:chain} with an additional estimable lower bound.

Based on the functional-quotient estimator, we propose a new diagnostic tool for understanding compression and redundancy in CPWL networks: The compression ratio
\begin{equation}\label{eq:rho_func}
	\rho_{\mathrm{func}}^{\,l}(\varepsilon)
	:= |\Omega_{\mathcal{D}, \mathrm{func}}^{\,l}(\varepsilon)| \,/\, |\Omega_{\mathcal{D}}^{\,l}|
	\;\in\; (0, 1]
\end{equation}
diagnoses how much pattern redundancy the network carries: small values indicate many distinct patterns collapsing onto few linear operators, values near $1$ indicate $\hat{I}_{\mathrm{func}} \approx \hat{I}(Y; \Omega_{\mathcal{D}})$. We track $\rho_{\mathrm{func}}^{\,l}(\varepsilon)$ as a layerwise diagnostic in \cref{sec:capacity-results}.

\section{Experimental setup}
\label{sec:experiments}

We evaluate the framework along two axes: \cref{sec:calibration-setup} \emph{calibrates} the plug-in routing estimator $\hat I(Y;\Omega_{\mathcal{D}})$ against three established MI baselines on three structurally distinct datasets, and \cref{sec:capacity-setup} \emph{stresses} the estimator at growing input dimension on PCA-reduced MNIST and tests whether the functional-equivalence quotient at relative tolerance $\varepsilon$ recovers informativeness when the raw partition saturates at $H(Y)$. For the baselines we use uniform binning ($K = 8$ bins per coordinate) and $k$-means clustering ($K = |\mathcal{Y}|$), both with the Miller--Madow correction, and the mixed continuous--discrete KSG estimator ($k = 3$) of \cite{rossMutualInformation2014}, the discrete-target adaptation of \cite{kraskovEstimatingMutualInformation2004}. Further, we use fully-connected ReLU MLPs trained for $150$ epochs with SGD with momentum \citep{sutskeverImportanceInitializationMomentum2013}, learning rate $\alpha = 10^{-3}$, momentum $\beta = 0.9$, batch size $32$, cross-entropy loss; no weight decay, dropout, or augmentation. Hidden weights use He initialisation \citep{heDelvingDeepRectifiers2015}, output weights Glorot initialisation \citep{glorotUnderstandingDifficultyTraining2010}, and biases start at zero. These settings were fixed in advance, not tuned. Five seeds per configuration; the data split is fixed across seeds. Networks are trained on $80\%$ of each dataset; every estimate, routing and baseline alike, is computed on the held-out $20\%$ (MNIST: the standard test split; \cref{app:train_test}). Implementation details, licences and compute are in \cref{app:implementation}.

\subsection{Study 1 -- Calibration against established estimators}
\label{sec:calibration-setup}

\paragraph{Datasets.}
\textit{Composite} -- a synthetic 7-class 2D mixture of moons, concentric circles, and Gaussian blobs ($10\,000$ points, of which $N = \res{heldout.composite.N_heldout}$ are held out for estimation; $H(Y) = \res{hy.composite.heldout}$ bits; \cref{app:composite_dataset}).
\textit{WBC} -- the UCI Wisconsin Breast Cancer tabular dataset \citep{williamwolbergBreastCancerWisconsin1993}, binary, $569$ points, of which $N = \res{heldout.wbc.N_heldout}$ are held out; $H(Y) = \res{hy.wbc.heldout}$ bits on the held-out points.
\textit{MNIST FC} -- MNIST~\cite{lecun1999mnist} flattened to $784$ dimensions, projected to $10$ PCA components, and rescaled to $[-1, 1]^{10}$; $N = 10\,000$-image test split, $H(Y) = \res{hy.mnist.heldout}$ bits.
The three datasets cover label entropies from $0.95$ to $3.32$ bits.

\paragraph{Architectures.}
Composite and WBC: six fully-connected ReLU architectures spanning depth $\in \{3, 5\}$ and width $\in \{5, 9, 25\}$. MNIST FC: three architectures $[3,3,3], [5,5,5], [7,7,7]$ -- wider or deeper networks enter the saturation regime studied in \cref{sec:capacity-setup}. $75$ networks in total.

\subsection{Study 2 -- Saturation and the functional-equivalence quotient}
\label{sec:capacity-setup}

We project MNIST onto its top $d$ principal components for $d \in \{2, 3, 4, 5, 10, 15, 20\}$, train three-hidden-layer ReLU MLPs with widths $\{7, 15, 25, 50\}$, and report estimators at the last hidden layer. The threshold $\varepsilon$ in \cref{eq:func_equiv} is evaluated on a grid of $15$ values, $\{0,\, 0.01,\, 0.05,\, 0.1,\, 0.2,\, \ldots,\, 1.0,\, 1.5,\, 2.0\}$; \cref{fig:mnist-capacity} shows the subset $\{0,\, 0.1,\, 0.2,\, 0.3,\, 0.4,\, 0.5,\, 1.0,\, 2.0\}$. This gives $7 \times 4 \times 5 = 140$ networks. The layerwise compression profile $\rho_{\mathrm{func}}^{\,l}(\varepsilon)$ is computed on the same grid for the five-layer Study-1 networks of Composite and WBC and for $[5,5,5]$ and $[7,7,7]$ on MNIST FC ($40$ networks). Training metrics are in \cref{app:train_metrics}.

\section{Results}
\label{sec:results}

\subsection{Calibration: pointwise agreement across networks}
\label{sec:calibration-scatter}

\begin{figure}[t]
	\centering
	\includegraphics[width=\linewidth]{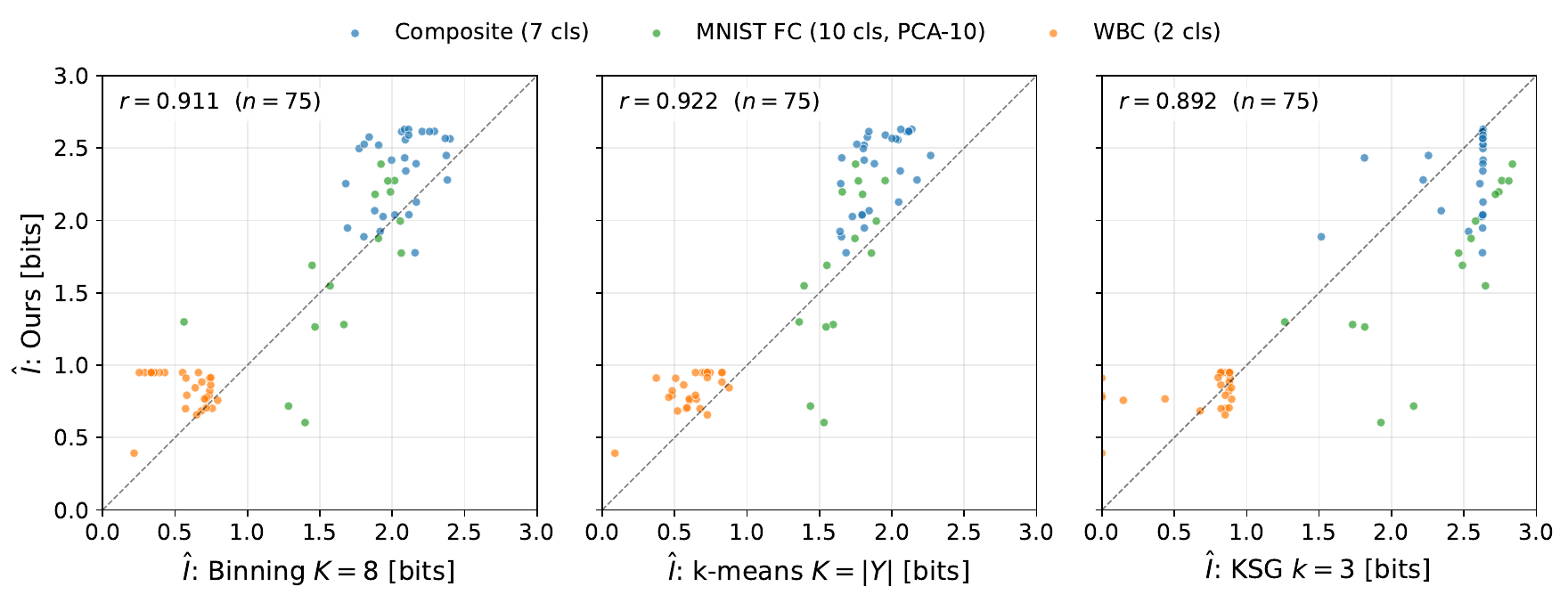}
	\caption{Pointwise calibration of $\hat I(Y;\Omega_{\mathcal{D}})$ against three MI baselines: binning and $k$-means with the Miller--Madow correction, and the $k$-nearest-neighbour estimator KSG (last epoch, last hidden layer, held-out points). There is one marker per network (architecture and seed) on \textit{Composite} (blue), \textit{WBC} (orange), and \textit{MNIST FC} (green). Pearson $r$ pooled across all three datasets: $r = \res{fig2.r.binning8}$ (binning, $K = 8$), $r = \res{fig2.r.kmeansKY}$ ($k$-means, $K = |\mathcal{Y}|$), $r = \res{fig2.r.ksg3}$ (KSG, $k = 3$). WBC networks with KSG near zero: \cref{app:ksg_ties}.}
	\label{fig:calibration-scatter}
\end{figure}

\Cref{fig:calibration-scatter} scatters $\hat I(Y;\Omega_{\mathcal{D}})$ against each of the three baselines, with one point per network: the last hidden layer at the last epoch, estimated on the same held-out points as the baselines ($n = \res{fig2.n}$ per panel). Within each dataset cluster the points lie along the diagonal: $\hat I$ assigns the same bit counts as the baselines without requiring a bin width, bandwidth, or cluster count. The drop in $r$ for KSG has two sources: (i) MNIST FC, where KSG sits systematically above $\hat I_{\mathrm{raw}}$ while binning and $k$-means track it closely, and (ii) Composite, where KSG sits at the $H(Y)$ ceiling for $\res{fig2.composite.ksg_within_0.02_of_H_Y}$ of the $\res{fig2.composite.n}$ networks (within $0.02$ bits of $H(Y) = \res{hy.composite.heldout}$) while $\hat I$ ranges from $\res{fig2.composite.routing_min}$ to $\res{fig2.composite.routing_max}$ bits.

\paragraph{Finite-sample regime.}
The estimator is trustworthy when most regions hold several points, i.e.\ at small occupancy $\rho$ (\cref{eq:rho}). Most calibration networks are in that regime, but the two width-$25$ WBC architectures are not: with $N = \res{heldout.wbc.N_heldout}$ held-out points they reach $\rho = \res{wbc_rho.[25, 25, 25].heldout}$ and $\res{wbc_rho.[25, 25, 25, 25, 25].heldout}$ at the deepest layer, so their points in \cref{fig:calibration-scatter} sit at $H(Y)$ by construction (\cref{app:occupancy}). Standard bias corrections do not repair a saturated partition (\cref{tab:bias_by_rho_compact}): Miller--Madow over-corrects by several bits, into negative values, while Grassberger and the coverage-based Chao--Shen and Chao--Wang--Jost corrections leave the estimate within $\res{beyond_mm.rho>0.9.gap_to_H_Y.chao_wang_jost}$ bits of $H(Y)$ at $\rho > 0.9$ (\cref{app:bias_corrections}). We therefore report $\rho$ with every estimate and treat saturation structurally (\cref{sec:capacity-results}).

\begin{table}[t]
	\centering
	\small
	\caption{Mean bias correction (plug-in minus corrected estimate, bits) by occupancy $\rho$, over the $\res{mm.n_cells}$ (dataset, architecture, PCA dimension, layer) cells of both studies; $n$: number of cells. Small corrections at $\rho > 0.9$ mean the corrected value stays at the $H(Y)$ ceiling. Full table: \cref{tab:bias_by_rho}.}
	\label{tab:bias_by_rho_compact}
\begin{tabular}{lrrr}
\toprule
Estimator & $\rho < 0.05$ ($n=41$) & $\rho > 0.3$ ($n=56$) & $\rho > 0.9$ ($n=19$) \\
\midrule
Miller--Madow & 0.114 & 3.739 & 5.403 \\
Grassberger & 0.029 & 0.067 & 0.006 \\
Chao--Shen & 0.032 & 0.139 & 0.054 \\
Chao--Wang--Jost & 0.032 & 0.185 & 0.065 \\
\bottomrule
\end{tabular}

\end{table}

\subsection{Calibration: layerwise profile at convergence}
\label{sec:calibration-layerwise}

\begin{figure}[t]
	\centering
	\includegraphics[width=\linewidth]{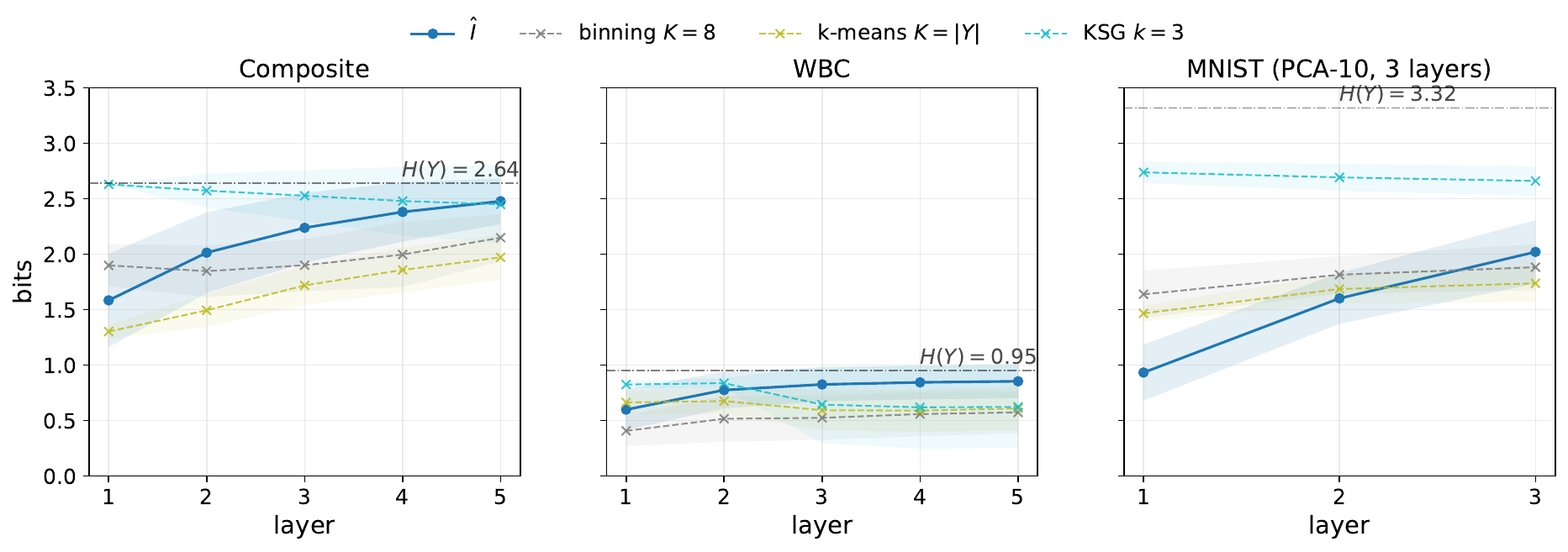}
	\caption{Layerwise bits at the last training epoch, mean $\pm 1\sigma$ across seeds and matching architectures. Solid: $\hat I(Y; \Omega_{\mathcal{D}})$. Dashed: the three baselines (binning and $k$-means with the Miller--Madow correction, KSG). Horizontal dash-dot: $H(Y)$. The Composite and WBC panels pool the three five-hidden-layer architectures; the MNIST FC panel pools width $5$ and $7$ networks.}
	\label{fig:layer-profile}
\end{figure}

On all three datasets in \cref{fig:layer-profile} $\hat I$ rises monotonically with depth, tracks binning and $k$-means at every layer, and stays at or below $H(Y)$ on every layer of every one of the $75$ networks (the chain~\cref{eq:chain} is empirically satisfied). KSG behaves differently: it starts near or above the routing estimator at layer $1$ and then \emph{drifts down} with depth on Composite ($\sim 2.6 \to 2.4$ bits), narrowing or inverting the KSG-to-routing gap by the deepest layer. On MNIST FC, KSG sits at $\sim 2.7$--$2.8$ bits across all three layers while $\hat I$ rises from $\sim 0.95$ to $\sim 2.0$, so the gap narrows from $\sim 1.8$ bits at layer $1$ to $\sim 0.65$ bits at layer $3$. Per \cref{cor:balance}, $I(Y;T) - I(Y;\Pi) = I(Y;T \mid \Pi) - I(Y;\Pi \mid T)$, so a positive KSG-above-routing gap signals that the within-region term $I(Y;T \mid \Pi)$ exceeds the routing-loss term $I(Y;\Pi \mid T)$: in narrow MNIST networks the continuous geometry $T$ within each region carries class information that the coarse routing partition $\Pi$ does not yet resolve, and the gap shrinks as deeper layers route more of that information into $\Pi$.

\subsection{Saturation and the functional-equivalence correction}
\label{sec:capacity-results}

\begin{figure}[t]
	\centering
	\includegraphics[width=\linewidth]{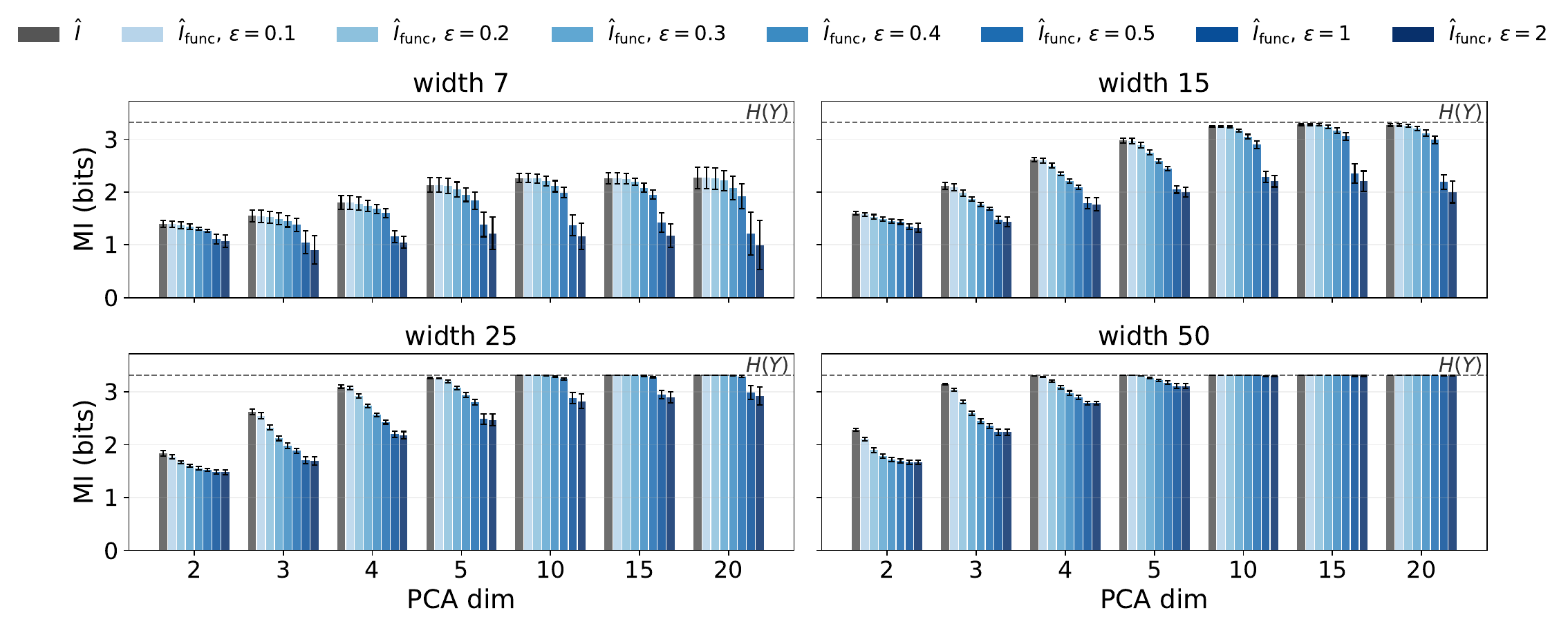}
	\caption{Routing MI on PCA-reduced MNIST (last layer, last epoch) as a function of input dimension $d$, for four 3-hidden-layer architectures. Bars: $\hat I$ (grey) and $\hat I_{\mathrm{func}}(\varepsilon)$ (light to dark blue); error bars $\pm 1\sigma$ over 5 seeds. Dashed line: $H(Y) = \res{capacity.H_Y}$. As $d$ grows the raw bar approaches $H(Y)$ for the wider architectures (saturation); the functional quotient at $\varepsilon \in [0.3, 0.5]$ pulls it back below $H(Y)$ except at widths $25$ and $50$.}
	\label{fig:mnist-capacity}
\end{figure}

\paragraph{Saturation.}
In \cref{fig:mnist-capacity}, for $d \leq 5$, $\hat I_{\mathrm{raw}}$ sits well below $H(Y)$ at widths $7$ and $15$ (by at least $\res{capacity.w15.d<=5.gap_min}$ bits): the partition is informative. Widths $25$ and $50$ approach $H(Y)$ already at small $d$ (up to $\res{capacity.w50.d<=5.raw_max}$ bits). As $d$ grows ($d \in \{10, 15, 20\}$) the raw estimator approaches $H(Y)$ for widths $25$ and $50$: nearly every sample gets its own region, so the empirical conditional entropy $\hat{H}(Y|\Omega_\mathcal{D})\to0$ and $\hat{I}_\mathrm{raw}\to\hat{H}(Y)$, whatever the network has learned.

\paragraph{Functional equivalence as correction.}
Merging starts at small $\varepsilon$: at $\varepsilon = 0.1$ and $0.2$, $\hat I_{\mathrm{func}}$ falls up to $\res{capacity.eps0.1.below_raw_max}$ and $\res{capacity.eps0.2.below_raw_max}$ bits below $\hat I_{\mathrm{raw}}$. At moderate $\varepsilon \in \{0.3, 0.4, 0.5\}$ the quotient restores informativeness only where there are redundant linear operators to collapse: at $d \geq 10$ and $\varepsilon = 0.5$ the width-$15$ estimate sits $\res{capacity.eps0.5.d>=10.w15.gap_min}$--$\res{capacity.eps0.5.d>=10.w15.gap_max}$ bits below $H(Y)$, the width-$25$ estimate only $\res{capacity.eps0.5.d>=10.w25.gap_min}$--$\res{capacity.eps0.5.d>=10.w25.gap_max}$ bits, and width $50$ stays saturated (gap $\leq \res{capacity.eps0.5.d>=10.w50.gap_max}$ bits): nearly every region there implements a distinct linear operator. At large $\varepsilon \in \{1, 2\}$ the quotient over-compresses; at $d = 2$, $\varepsilon = 2$ keeps $\res{capacity.eps2.d2.func_over_raw_min}$--$\res{capacity.eps2.d2.func_over_raw_max}$ of the raw value.

\begin{figure}[t]
	\centering
	\includegraphics[width=\linewidth]{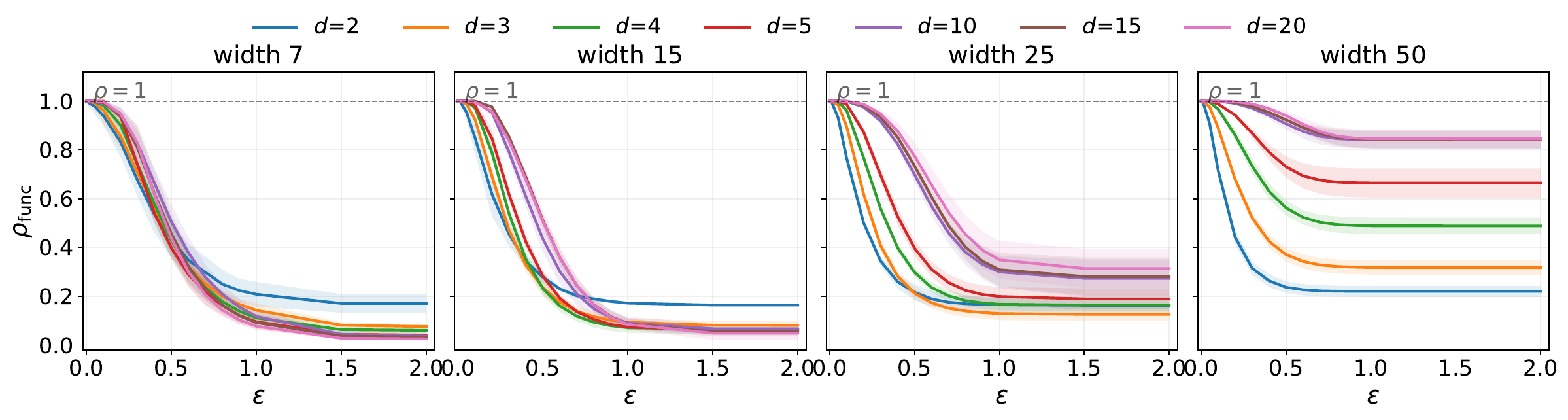}
	\caption{$\rho_{\mathrm{func}}(\varepsilon)$ on PCA-reduced MNIST of dimension $d$ at the last hidden layer and training epoch. Lower $\rho_{\mathrm{func}}$ means more aggressive merging.}
	\label{fig:mnist-rho-eps}
\end{figure}

\paragraph{Compression-ratio diagnostic and layerwise compression.}
\Cref{fig:mnist-rho-eps} traces $\rho_{\mathrm{func}}(\varepsilon)$ at the last hidden layer. Small-$d$ networks ($d{=}2, 3$) drop sharply at $\varepsilon = 0.1$ -- many distinct routing paths share nearly identical linear operators -- while large-$d$ networks ($d{=}15, 20$) stay near $1$ until $\varepsilon \approx 0.2$--$0.3$ before dropping. The transition window aligns with the window in \cref{fig:mnist-capacity} where $\hat I_{\mathrm{func}}$ provides informative compression: both diagnostics identify $\varepsilon \in [0.2, 0.5]$ as the operating range. Across layers of the calibration networks (\cref{fig:rho-func-layerwise}, \cref{app:func_layerwise}), $\rho_{\mathrm{func}}^{\,1} = 1$ for every network, and for $\varepsilon > 0$ the ratio decreases with depth on average: functional equivalence is a phenomenon of compositions.

\section{Discussion}
\label{sec:discussion}

\subsection{Calibration against established MI estimators}
\label{sec:calibration-discussion}

\paragraph{What the calibration tests.}
$\hat I(Y; \Omega_{\mathcal{D}})$ is not proposed as a proxy for $I(Y;T)$: by \cref{cor:balance} the two can differ in either direction, and the KSG--routing gap discussed below shows that they do. The calibration in \cref{sec:calibration-scatter} is an internal-consistency check of the plug-in estimator: applied to the same networks and data, independent discretisation schemes should recover comparable information, and they do. Agreement between finite-sample estimators cannot rule out a bias they share, and the occupancy $\rho$ (\cref{eq:rho}) marks where that matters. The two width-$25$ WBC architectures sit at $\rho = \res{wbc_rho.[25, 25, 25].heldout}$ and $\res{wbc_rho.[25, 25, 25, 25, 25].heldout}$, where every estimate is pushed towards $H(Y)$ and agreement carries little information (\cref{app:occupancy}).

\paragraph{Why the routing estimator tracks the discrete baselines.}
The Pearson coefficients $r = \res{fig2.r.binning8}$ (binning) and $r = \res{fig2.r.kmeansKY}$ ($k$-means) over $\res{fig2.n}$ networks place $\hat I(Y; \Omega_{\mathcal{D}})$ within the intrinsic plug-in bias of the discrete baselines, across three structurally distinct datasets and a range from $0$ bits up to $H(Y)$. The structural distinction is that $\Omega_{\mathcal{D}}$ is read from the network's piecewise-linear geometry rather than from an external discretisation: no bin width, bandwidth, or cluster count is tuned, and the chain~\cref{eq:chain} fixes what the estimator measures (the routing information $I(Y;\Pi)$, lower-bounding $I(Y;\Pi, T)$).

\paragraph{The KSG--routing gap localises within-region information.}
KSG estimates $I(Y; T)$ from the continuous representation while $\hat I(Y;\Omega_{\mathcal{D}})$ estimates $I(Y;\Pi)$, so by \cref{cor:balance} a positive KSG-above-routing gap reads as $I(Y; T \mid \Pi) > I(Y; \Pi \mid T)$: the within-region geometry $T \mid \Pi$ carries class information that the partition $\Pi$ has not yet routed. The narrowing of the gap with depth on all three datasets, and the inversion on Composite where the routing estimator overtakes KSG by layer~$5$, tracks $\Pi$ becoming more class-discriminative. KSG's known overestimation in low-dimensional continuous representations \citep{kraskovEstimatingMutualInformation2004,goldfeldEstimatingInformationFlow2019} likely contributes some additional gap on MNIST FC; disentangling the two contributions requires an independent estimator of $I(Y; T \mid \Pi)$.

\paragraph{The $H(Y)$ ceiling holds.}
Across all $75$ networks, every hidden layer, and four estimators, no estimate exceeds $H(Y)$ in \cref{fig:layer-profile}; the chain~\cref{eq:chain} is empirically tight on MNIST FC and WBC and approached on Composite, across an order of magnitude of label entropy.

\subsection{Functional equivalence as both correction and diagnostic}
\label{sec:rho-discussion}

\paragraph{Saturation correction.}
When the input dimension is large enough that $|\Omega_{\mathcal{D}}|$ approaches $N$, almost every region is a singleton, $\hat{I}_\mathrm{raw}\approx\hat{H}(Y)$, and the chain~\cref{eq:chain} carries no information. The functional quotient at $\varepsilon \in [0.3, 0.5]$ collapses routing paths that implement nearly identical linear operators and pulls the estimate back below $H(Y)$ when the network has such redundancy (widths $7$--$15$ in \cref{fig:mnist-capacity}); it cannot when nearly every region implements a distinct operator (width $50$). Inside the window the estimate still moves with $\varepsilon$, by up to $\res{capacity.window.max_abs_diff_eps0.3_eps0.5}$ bits between $\varepsilon = 0.3$ and $0.5$. The criterion \cref{eq:func_equiv} is bounded in $[0, 2]$, dimensionless, and depth- and architecture-invariant, so $\varepsilon$ is a reportable choice rather than a tuned hyperparameter; \cref{fig:mnist-rho-eps} provides an audit of whether the chosen $\varepsilon$ is in the merging window.

\paragraph{$\rho_{\mathrm{func}}^{\,l}(\varepsilon)$ as a structural fingerprint.}
Two structural facts shape \cref{fig:rho-func-layerwise}: $\rho_{\mathrm{func}}^{\,1} = 1$ identically (a single ReLU layer maps every activation pattern to a distinct linear operator), and on average $\rho_{\mathrm{func}}^{\,l}$ decreases in $l$ at any fixed $\varepsilon > 0$ (each composition opens new opportunities for products of selection matrices to coincide). The dataset ordering -- Composite most compressible, MNIST FC PCA-10 most resistant at small $\varepsilon$ -- reads as how much pattern redundancy each network develops when the data supports it: Composite admits a compact functional description by layer~$5$, while MNIST FC PCA-10, already in the saturation regime, requires non-trivial $\varepsilon$ before any merging.
$\rho_{\mathrm{func}} \approx 1$ throughout means $\hat I_{\mathrm{func}} \approx \hat I_{\mathrm{raw}}$ and the quotient is not worth computing; large drops signal that it is. On MNIST FC the routing and functional estimates separate as $\rho_{\mathrm{func}}$ drops, as the diagnostic predicts.

\paragraph{Limitations and scope} The MNIST FC calibration was obtained on networks narrow enough that $|\Omega_{\mathcal{D}}|$ remains below $N$. For wider or deeper networks the raw estimator is pushed by $H(Y)$, and the functional quotient is required (Study~2). Moreover, the plug-in estimator \cref{eq:plugin} is bounded away from $I(Y;\Pi)$ by $H(Y)\,\Pr(\Pi\notin\Omega_{\mathcal{D}}) + h_2(\Pr(\Pi\notin\Omega_{\mathcal{D}}))$ (\cref{app:proof_routing_est}). The second term dominates near the saturation threshold and surfaces in \cref{fig:mnist-capacity} as the closing gap between $\hat I_{\mathrm{raw}}$ and $H(Y)$ at large $d$. A bound that tracks the occupancy $\rho$ explicitly would sharpen this; until then we report $\rho$ with every estimate, and by that measure the two width-$25$ WBC calibration architectures ($\rho \geq \res{wbc_rho.[25, 25, 25].heldout}$) lie outside the reliable regime (\cref{sec:calibration-scatter}). High routing information is also not by itself evidence of learned class structure. With randomly permuted training labels (\cref{app:label_permutation}), the narrow $[5,5,5]$ Composite network's information about the true labels falls from $\res{permutation.composite.[5, 5, 5].clean.raw}$ to $\res{permutation.composite.[5, 5, 5].permuted.raw}$ bits, and the functional quotient discounts it further ($\res{permutation.composite.[5, 5, 5].clean.quotient}$ to $\res{permutation.composite.[5, 5, 5].permuted.quotient}$ bits). The wide $[25,25,25]$ network keeps $\res{permutation.composite.[25, 25, 25].permuted.raw}$ bits raw and $\res{permutation.composite.[25, 25, 25].permuted.quotient}$ bits after the quotient, close to $H(Y) = \res{hy.composite.heldout}$ bits, at $\res{permutation.composite.[25, 25, 25].permuted.test_accuracy.pct}\%$ true-label test accuracy: a fine partition of spatially coherent classes correlates with the class whether or not the network has learned it. The quotient remains a valid lower bound by the data-processing inequality, but it removes this effect only at moderate capacity, so a high estimate should be read together with $\rho$ and, where possible, a label-permutation control. Lastly, the quotient estimate depends on $\varepsilon$ even inside the window $[0.2, 0.5]$: small $\varepsilon$ already merges some regions ($\rho_{\mathrm{func}}(0.1) \geq \res{capacity.eps0.1.rho_func_min}$) and $\varepsilon \geq 1$ over-merges. We report $\varepsilon$ with every estimate and audit it through $\rho_{\mathrm{func}}$ rather than treat it as an opaque hyperparameter.

\section{Conclusion and Outlook}
\label{sec:conclusion}
We introduced a framework for the internal information geometry of CPWL networks: an exact decomposition of $I(Y;T)$ into routing, within-region, and routing-loss terms; a hyperparameter-free plug-in routing estimator $\hat I(Y;\Omega_{\mathcal{D}})$ from the piecewise-linear partition; and a functional-equivalence quotient extending the estimator into the fine-resolution regime, at the cost of one bounded parameter $\varepsilon$ that is reported and audited through $\rho_{\mathrm{func}}(\varepsilon)$. As an estimator sanity check, $\hat I$ agrees with binning and $k$-means ($r = \res{fig2.r.binning8}$, $\res{fig2.r.kmeansKY}$) and remains at or below $H(Y)$ across all 75 networks; the KSG-above-routing gap localises within-region information rather than estimator disagreement; and the functional quotient recovers informative estimates on PCA-reduced MNIST where the raw estimator saturates at $H(Y)$. Future work includes estimating the decomposition gap directly and extending the analysis to convolutional and residual architectures, where nearly every data point occupies its own region.


\begin{ack}
	The authors would like to thank associate professor Marta Panizzut at UiT - The Arctic University of Norway for fruitful discussions on algebraic geometry and tropical algebra.

	This work was supported by the Research Council of Norway through its Centre of Excellence Integreat -- The Norwegian Centre for knowledge-driven machine learning, project number 332645.

	Ali Ramezani-Kebrya was supported by the Research Council of Norway through FRIPRO Grant
	under project number 356103.
\end{ack}

\newpage
\bibliography{geobin_final}

@inproceedings{adilovaInformationPlaneAnalysis2022,
  title = {Information Plane Analysis for Dropout Neural Networks},
  author = {Adilova, Linara and Geiger, Bernhard C. and Fischer, Asja},
  booktitle = {International Conference on Learning Representations},
  year = {2023},
}

@inproceedings{belghaziMutualInformationNeural2018,
  title = {Mutual Information Neural Estimation},
  author = {Belghazi, Mohamed Ishmael and Baratin, Aristide and Rajeshwar, Sai
            and Ozair, Sherjil and Bengio, Yoshua and Courville, Aaron and Hjelm,
            Devon},
  booktitle = {Proceedings of the 35th International Conference on Machine
               Learning},
  series = {PMLR},
  volume = {80},
  pages = {531--540},
  year = {2018},
}

@inproceedings{berzinsPolyhedralComplexExtraction2023,
  title = {Polyhedral Complex Extraction from {ReLU} Networks Using Edge
           Subdivision},
  author = {Berzins, Arturs},
  booktitle = {Proceedings of the 40th International Conference on Machine
               Learning},
  series = {PMLR},
  volume = {202},
  pages = {2234--2244},
  year = {2023},
}

@misc{dhayalkarGeometryReLUNetworks2025,
  title = {The Geometry of {ReLU} Networks through the {ReLU} Transition Graph},
  author = {Dhayalkar, Sahil Rajesh},
  howpublished = {arXiv preprint arXiv:2505.11692},
  year = {2025},
}

@INPROCEEDINGS{dhayalkar2025discretefunctionalgeometryrelu,
  author={Dhayalkar, Sahil Rajesh},
  booktitle={2025 5th International Conference on Robotics, Automation, and Artificial Intelligence (RAAI)}, 
  title={Discrete Functional Geometry of ReLU Networks via ReLU Transition Graphs}, 
  year={2025},
  volume={},
  number={},
  pages={153-159},
  doi={10.1109/RAAI67517.2025.11423317}}

@inproceedings{felUnderstandingVisualFeature2024b,
  title = {Understanding Visual Feature Reliance through the Lens of Complexity},
  author = {Fel, Thomas and B{\'e}thune, Louis and Lampinen, Andrew K. and Serre
            , Thomas and Hermann, Katherine},
  booktitle = {Advances in Neural Information Processing Systems},
  volume = {37},
  pages = {69888--69924},
  year = {2024},
}

@article{geigerInformationPlaneAnalyses2022,
  title = {On Information Plane Analyses of Neural Network Classifiers -- {A}
           Review},
  author = {Geiger, Bernhard C.},
  journal = {IEEE Transactions on Neural Networks and Learning Systems},
  volume = {33},
  number = {12},
  pages = {7039--7051},
  year = {2022},
  doi = {10.1109/TNNLS.2021.3089037},
}

@inproceedings{glorotDeepSparseRectifier2011,
  title = {Deep Sparse Rectifier Neural Networks},
  author = {Glorot, Xavier and Bordes, Antoine and Bengio, Yoshua},
  booktitle = {Proceedings of the 14th International Conference on Artificial
               Intelligence and Statistics},
  series = {PMLR},
  volume = {15},
  pages = {315--323},
  year = {2011},
}

@inproceedings{goldfeldEstimatingInformationFlow2019,
  title = {Estimating Information Flow in Deep Neural Networks},
  author = {Goldfeld, Ziv and van den Berg, Ewout and Greenewald, Kristjan and
            Melnyk, Igor and Nguyen, Nam and Kingsbury, Brian and Polyanskiy,
            Yury},
  booktitle = {Proceedings of the 36th International Conference on Machine
               Learning},
  series = {PMLR},
  volume = {97},
  pages = {2299--2308},
  year = {2019},
}

@inproceedings{goldfeldSlicedMutualInformation2021,
  title = {Sliced Mutual Information: A Scalable Measure of Statistical
           Dependence},
  author = {Goldfeld, Ziv and Greenewald, Kristjan},
  booktitle = {Advances in Neural Information Processing Systems},
  volume = {34},
  pages = {17567--17578},
  year = {2021},
}

@article{goujonNumberRegionsPiecewise2023,
  title = {On the Number of Regions of Piecewise Linear Neural Networks},
  author = {Goujon, Alexis and Etemadi, Arian and Unser, Michael},
  journal = {Journal of Computational and Applied Mathematics},
  volume = {441},
  pages = {115667},
  year = {2024},
  doi = {10.1016/j.cam.2023.115667},
}

@inproceedings{haninComplexityLinearRegions2019,
  title = {Complexity of Linear Regions in Deep Networks},
  author = {Hanin, Boris and Rolnick, David},
  booktitle = {Proceedings of the 36th International Conference on Machine
               Learning},
  series = {PMLR},
  volume = {97},
  pages = {2596--2604},
  year = {2019},
}

@inproceedings{haninDeepReLUNetworks2019,
  title = {Deep {ReLU} Networks Have Surprisingly Few Activation Patterns},
  author = {Hanin, Boris and Rolnick, David},
  booktitle = {Advances in Neural Information Processing Systems},
  volume = {32},
  year = {2019},
}

@inproceedings{huInfoNetNeuralEstimation2024a,
  title = {{InfoNet}: Neural Estimation of Mutual Information without Test-Time
           Optimization},
  author = {Hu, Zhengyang and Kang, Song and Zeng, Qunsong and Huang, Kaibin and
            Yang, Yanchao},
  booktitle = {Proceedings of the 41st International Conference on Machine
               Learning},
  series = {PMLR},
  volume = {235},
  pages = {19283--19303},
  year = {2024},
}

@inproceedings{humayunSplineCamExactVisualization2024,
  title = {{SplineCam}: Exact Visualization and Characterization of Deep Network
           Geometry and Decision Boundaries},
  author = {Humayun, Ahmed Imtiaz and Balestriero, Randall and Balakrishnan,
            Guha and Baraniuk, Richard},
  booktitle = {Proceedings of the IEEE/CVF Conference on Computer Vision and
               Pattern Recognition},
  pages = {3789--3798},
  year = {2023},
}

@inproceedings{jeonInformationTheoreticDiscreteDiffusion2025,
  title = {Information-Theoretic Discrete Diffusion},
  author = {Jeon, Moongyu and Shin, Sangwoo and Jeon, Dongjae and No, Albert},
  booktitle = {Advances in Neural Information Processing Systems},
  volume = {38},
  pages = {20913--20939},
  year = {2025},
}

@article{kraskovEstimatingMutualInformation2004,
  title = {Estimating Mutual Information},
  author = {Kraskov, Alexander and St{\"o}gbauer, Harald and Grassberger, Peter},
  journal = {Physical Review E},
  volume = {69},
  number = {6},
  pages = {066138},
  year = {2004},
  doi = {10.1103/PhysRevE.69.066138},
}

@article{lecun1999mnist,
  title = {Gradient-Based Learning Applied to Document Recognition},
  author = {LeCun, Yann and Bottou, L{\'e}on and Bengio, Yoshua and Haffner,
            Patrick},
  journal = {Proceedings of the IEEE},
  volume = {86},
  number = {11},
  pages = {2278--2324},
  year = {1998},
  doi = {10.1109/5.726791},
}

@inproceedings{lewandowskiExploitingSpaceFolding2026,
  title = {Exploiting Space Folding by Neural Networks},
  author = {Lewandowski, Michal and Pisoni, Raphael and Heinzl, Bernhard and
            Moser, Bernhard A.},
  booktitle = {Proceedings of the AAAI Conference on Artificial Intelligence},
  volume = {40},
  pages = {22833--22840},
  year = {2026},
  doi = {10.1609/aaai.v40i27.39446},
}

@article{liuIntegratingGeometriesReLU2023a,
  title = {Integrating Geometries of {ReLU} Feedforward Neural Networks},
  author = {Liu, Yajing and Caglar, Turgay and Peterson, Christopher and Kirby,
            Michael},
  journal = {Frontiers in Big Data},
  volume = {6},
  pages = {1274831},
  year = {2023},
  doi = {10.3389/fdata.2023.1274831},
}

@incollection{miller1955note,
  title = {Note on the Bias of Information Estimates},
  booktitle = {Information Theory in Psychology: {{Problems}} and Methods},
  author = {Miller, George A.},
  editor = {Quastler, Henry},
  year = 1955,
  pages = {95--100},
  publisher = {Free Press},
  address = {Glencoe, IL},
}

@article{moonEstimationMutualInformation1995,
  title = {Estimation of Mutual Information Using Kernel Density Estimators},
  author = {Moon, Young-Il and Rajagopalan, Balaji and Lall, Upmanu},
  journal = {Physical Review E},
  volume = {52},
  number = {3},
  pages = {2318--2321},
  year = {1995},
  doi = {10.1103/PhysRevE.52.2318},
}

@inproceedings{NIPS2014_fa6f2a46,
  title = {On the Number of Linear Regions of Deep Neural Networks},
  author = {Mont{\'u}far, Guido and Pascanu, Razvan and Cho, Kyunghyun and
            Bengio, Yoshua},
  booktitle = {Advances in Neural Information Processing Systems},
  volume = {27},
  year = {2014},
}

@misc{oordRepresentationLearningContrastive2018,
  title = {Representation Learning with Contrastive Predictive Coding},
  author = {van den Oord, A{\"a}ron and Li, Yazhe and Vinyals, Oriol},
  howpublished = {arXiv preprint arXiv:1807.03748},
  year = {2018},
}

@article{paninskiEstimationEntropyMutual2003a,
  title = {Estimation of Entropy and Mutual Information},
  author = {Paninski, Liam},
  journal = {Neural Computation},
  volume = {15},
  number = {6},
  pages = {1191--1253},
  year = {2003},
  doi = {10.1162/089976603321780272},
}

@inproceedings{parkUnsupervisedRepresentationLearning,
  title = {Unsupervised Representation Learning via Neural Activation Coding},
  author = {Park, Yookoon and Lee, Sangho and Kim, Gunhee and Blei, David},
  booktitle = {Proceedings of the 38th International Conference on Machine
               Learning},
  series = {PMLR},
  volume = {139},
  pages = {8391--8400},
  year = {2021},
}

@misc{pascanuNumberResponseRegions2013,
  title = {On the Number of Response Regions of Deep Feed Forward Networks with
           Piece-Wise Linear Activations},
  author = {Pascanu, Razvan and Mont{\'u}far, Guido and Bengio, Yoshua},
  howpublished = {arXiv preprint arXiv:1312.6098},
  year = {2013},
}

@article{rossMutualInformation2014,
  title = {Mutual Information between Discrete and Continuous Data Sets},
  author = {Ross, Brian C.},
  journal = {PLoS ONE},
  volume = {9},
  number = {2},
  pages = {e87357},
  year = {2014},
  doi = {10.1371/journal.pone.0087357},
}

@article{sattelbergLocallyLinearAttributes2023,
  title = {Locally Linear Attributes of {ReLU} Neural Networks},
  author = {Sattelberg, Ben and Cavalieri, Renzo and Kirby, Michael and Peterson
            , Chris and Beveridge, Ross},
  journal = {Frontiers in Artificial Intelligence},
  volume = {6},
  pages = {1255192},
  year = {2023},
  doi = {10.3389/frai.2023.1255192},
}

@article{saxeInformationBottleneckTheory2019,
  title = {On the Information Bottleneck Theory of Deep Learning},
  author = {Saxe, Andrew M. and Bansal, Yamini and Dapello, Joel and Advani,
            Madhu and Kolchinsky, Artemy and Tracey, Brendan D. and Cox, David D.
            },
  journal = {Journal of Statistical Mechanics: Theory and Experiment},
  volume = {2019},
  number = {12},
  pages = {124020},
  year = {2019},
  doi = {10.1088/1742-5468/ab3985},
}

@inproceedings{serraBoundingCountingLinear2018,
  title = {Bounding and Counting Linear Regions of Deep Neural Networks},
  author = {Serra, Thiago and Tjandraatmadja, Christian and Ramalingam, Srikumar
            },
  booktitle = {Proceedings of the 35th International Conference on Machine
               Learning},
  series = {PMLR},
  volume = {80},
  pages = {4558--4566},
  year = {2018},
}

@misc{shwartz-zivOpeningBlackBox2017,
  title = {Opening the Black Box of Deep Neural Networks via Information},
  author = {{Shwartz-Ziv}, Ravid and Tishby, Naftali},
  howpublished = {arXiv preprint arXiv:1703.00810},
  year = {2017},
}

@inproceedings{tishbyDeepLearningInformation2015a,
  title = {Deep Learning and the Information Bottleneck Principle},
  author = {Tishby, Naftali and Zaslavsky, Noga},
  booktitle = {IEEE Information Theory Workshop (ITW)},
  pages = {1--5},
  year = {2015},
  doi = {10.1109/ITW.2015.7133169},
}

@inproceedings{tishbyInformationBottleneckMethod2000a,
  title = {The Information Bottleneck Method},
  author = {Tishby, Naftali and Pereira, Fernando C. and Bialek, William},
  booktitle = {Proceedings of the 37th Annual Allerton Conference on
               Communication, Control, and Computing},
  pages = {368--377},
  year = {1999},
}

@inproceedings{vincentReachablePolyhedralMarching2021,
  title = {Reachable Polyhedral Marching ({RPM}): A Safety Verification
           Algorithm for Robotic Systems with Deep Neural Network Components},
  author = {Vincent, Joseph A. and Schwager, Mac},
  booktitle = {IEEE International Conference on Robotics and Automation (ICRA)},
  pages = {9029--9035},
  year = {2021},
  doi = {10.1109/ICRA48506.2021.9561956},
}

@misc{williamwolbergBreastCancerWisconsin1993,
  title = {Breast Cancer {Wisconsin} (Diagnostic)},
  author = {Wolberg, William H. and Street, W. Nick and Mangasarian, Olvi L.},
  howpublished = {UCI Machine Learning Repository},
  year = {1993},
  doi = {10.24432/C5DW2B},
}

@inproceedings{wongsoUsingSlicedMutual2023,
  title = {Using Sliced Mutual Information to Study Memorization and
           Generalization in Deep Neural Networks},
  author = {Wongso, Shelvia and Ghosh, Rohan and Motani, Mehul},
  booktitle = {Proceedings of the 26th International Conference on Artificial
               Intelligence and Statistics},
  series = {PMLR},
  volume = {206},
  pages = {11608--11629},
  year = {2023},
}

@inproceedings{gaines2026characterizing,
  title = {Characterizing the Discrete Geometry of {ReLU} Networks},
  author = {Gaines, Blake B. and Bi, Jinbo},
  booktitle = {International Conference on Learning Representations},
  year = {2026},
  url = {https://openreview.net/forum?id=TgLW2DiRDG},
}

@article{buthInfomeasureComprehensivePython2025,
  title = {infomeasure: A Comprehensive {Python} Package for Information Theory
           Measures and Estimators},
  author = {B{\"u}th, Carlson Moses and Acharya, Kishor and Zanin, Massimiliano},
  journal = {Scientific Reports},
  volume = {15},
  number = {1},
  pages = {29323},
  year = {2025},
  doi = {10.1038/s41598-025-14053-5},
}

@misc{grassbergerEntropyEstimatesInsufficient2003,
  title = {Entropy Estimates from Insufficient Samplings},
  author = {Grassberger, Peter},
  howpublished = {arXiv preprint arXiv:physics/0307138},
  year = {2003},
}

@article{chaoNonparametricEstimationShannon2003,
  title = {Nonparametric Estimation of {Shannon}'s Index of Diversity When There
           Are Unseen Species in Sample},
  author = {Chao, Anne and Shen, Tsung-Jen},
  journal = {Environmental and Ecological Statistics},
  volume = {10},
  number = {4},
  pages = {429--443},
  year = {2003},
  doi = {10.1023/A:1026096204727},
}

@article{chaoEntropySpeciesAccumulation2013,
  title = {Entropy and the Species Accumulation Curve: A Novel Entropy Estimator
           via Discovery Rates of New Species},
  author = {Chao, Anne and Wang, Y. T. and Jost, Lou},
  journal = {Methods in Ecology and Evolution},
  volume = {4},
  number = {11},
  pages = {1091--1100},
  year = {2013},
  doi = {10.1111/2041-210X.12108},
}

@article{nemenmanCoincidencesEstimationEntropies2011,
  title = {Coincidences and Estimation of Entropies of Random Variables with
           Large Cardinalities},
  author = {Nemenman, Ilya},
  journal = {Entropy},
  volume = {13},
  number = {12},
  pages = {2013--2023},
  year = {2011},
  doi = {10.3390/e13122013},
}

@inproceedings{sutskeverImportanceInitializationMomentum2013,
  title = {On the Importance of Initialization and Momentum in Deep Learning},
  author = {Sutskever, Ilya and Martens, James and Dahl, George and Hinton,
            Geoffrey},
  booktitle = {Proceedings of the 30th International Conference on Machine
               Learning},
  series = {PMLR},
  volume = {28},
  pages = {1139--1147},
  year = {2013},
}

@inproceedings{heDelvingDeepRectifiers2015,
  title = {Delving Deep into Rectifiers: Surpassing Human-Level Performance on {
           ImageNet} Classification},
  author = {He, Kaiming and Zhang, Xiangyu and Ren, Shaoqing and Sun, Jian},
  booktitle = {Proceedings of the IEEE International Conference on Computer
               Vision},
  pages = {1026--1034},
  year = {2015},
}

@inproceedings{glorotUnderstandingDifficultyTraining2010,
  title = {Understanding the Difficulty of Training Deep Feedforward Neural
           Networks},
  author = {Glorot, Xavier and Bengio, Yoshua},
  booktitle = {Proceedings of the 13th International Conference on Artificial
               Intelligence and Statistics},
  series = {PMLR},
  volume = {9},
  pages = {249--256},
  year = {2010},
}

@inproceedings{paszkePyTorchImperativeStyle2019,
  title = {{PyTorch}: An Imperative Style, High-Performance Deep Learning
           Library},
  author = {Paszke, Adam and Gross, Sam and Massa, Francisco and Lerer, Adam and
            Bradbury, James and Chanan, Gregory and Killeen, Trevor and Lin,
            Zeming and Gimelshein, Natalia and Antiga, Luca and others},
  booktitle = {Advances in Neural Information Processing Systems},
  volume = {32},
  year = {2019},
}

@article{pedregosaScikitlearnMachineLearning2011,
  title = {Scikit-learn: Machine Learning in {Python}},
  author = {Pedregosa, Fabian and Varoquaux, Ga{\"e}l and Gramfort, Alexandre
            and Michel, Vincent and Thirion, Bertrand and Grisel, Olivier and
            Blondel, Mathieu and Prettenhofer, Peter and Weiss, Ron and Dubourg,
            Vincent and others},
  journal = {Journal of Machine Learning Research},
  volume = {12},
  pages = {2825--2830},
  year = {2011},
}

@software{parx,
author = {Mylius-Kroken, Johan},
doi = {10.5281/zenodo.22696450},
license = {MIT},
title = {{parx: Polyhedral Affine Region eXplorer}},
url = {https://github.com/Johanmkr/parx}
}

\newpage
\onecolumn
\makeatletter
\let\addcontentsline\origaddcontentsline
\makeatother

\etocsettocstyle{\section*{Table of Contents}}{}
\etocsetnexttocdepth{subsection}
\etoctableofcontents

\newpage
\appendix
\newpage
\appendix
\crefalias{section}{appendix}
\crefalias{subsection}{appendix}

\section{Detailed Formulation of Linear Regions for CPWL Networks}
\label{app:linear_regions}

    \paragraph{Notation.} This appendix separates the contribution of a
    single layer from the partition through that layer. We write $\Pi^l$ for
    the partition induced by the hyperplanes of layer $l$ alone and
    $\Pi^{\leq l}$ for the common refinement through layer $l$; the latter
    is the partition written $\Pi^l$ in the main text. As there,
    $\pi^i_\omega$ is the activation pattern of region $\omega$ at layer $i$,
    and $\Omega$ is reserved for data-supported subsets
    $\Omega_{\mathcal{D}}^{\leq l} \subset \Pi^{\leq l}$. Matrices are
    written without boldface. The symbol $S^\omega_i$
    (\cref{subsec:functional_quotient}) denotes the set of active
    neurons at layer $i$ in region $\omega$; to avoid collision we use
    $\boldsymbol{\sigma}^l_\omega, \Sigma^l_\omega$ for the sign vector and
    sign diagonal introduced below.

    Let a CPWL network $\mathcal{N}: \mathcal{X} \to \mathbb{R}^k$ with $L$ 
    layers be expressed as
    \begin{equation}
        \mathcal{N}(\mathbf{x}) = F^L \circ F^{L-1} \circ \cdots \circ F^1(\mathbf{x}),
        \qquad 
        F^l(\mathbf{z}) = \sigma(W^l \mathbf{z} + \mathbf{b}^l),
    \end{equation}
    with weights $W^l$, biases $\mathbf{b}^l$, and piecewise-linear activation 
    $\sigma$. Each layer induces a partition $\Pi^l$ of the input space into 
    convex linear regions; the arrangement through layer $l$ is the common 
    refinement $\Pi^{\leq l} = \bigcap_{j=1}^{l} \Pi^j$.

    \paragraph{Per-region affine form.}
    Within $\omega \in \Pi^{\leq l}$ the restriction $\mathcal{N}^l|_\omega$ is 
    affine: $\mathcal{N}^l(\mathbf{x}) = A^l_\omega \mathbf{x} + \mathbf{c}^l_\omega$ 
    with
    \begin{equation}
        A^l_\omega = \prod_{i=l}^{1} Q^i_\omega W^i,
        \qquad 
        \mathbf{c}^l_\omega = Q^l_\omega \mathbf{b}^l 
            + \sum_{j=1}^{l-1} 
              \left(\prod_{k=l}^{j+1} Q^k_\omega W^k\right) Q^j_\omega \mathbf{b}^j,
    \end{equation}
    where the descending product convention $\prod_{i=l}^{1} M_i := M_l M_{l-1} \cdots M_1$ 
    matches feed-forward composition, and $Q^i_\omega = \mathrm{diag}(\pi^i_\omega)$ 
    is the diagonal matrix that selects active neurons at layer $i$ for region 
    $\omega$\footnote{For ReLU,
    $\pi^i_\omega = \mathbf{1}[\mathbf{z}^i > 0]$ is the indicator of positive
    pre-activations at layer $i$, as in the main text; the activation patterns
    for $i < l$ are determined by the inequalities defining $\omega$.}. Equivalently, $A^l_\omega$ and 
    $\mathbf{c}^l_\omega$ satisfy the recursion
    \begin{equation}
        A^l_\omega = Q^l_\omega W^l A^{l-1}_\omega,
        \qquad 
        \mathbf{c}^l_\omega = Q^l_\omega(W^l \mathbf{c}^{l-1}_\omega + \mathbf{b}^l),
        \qquad 
        A^0_\omega = I, \; \mathbf{c}^0_\omega = \mathbf{0}.
    \end{equation}
    The quantities $W^l A^{l-1}_\omega$ and 
    $W^l \mathbf{c}^{l-1}_\omega + \mathbf{b}^l$ are the effective weights and 
    biases of layer $l$ restricted to region $\omega$.

    \paragraph{Projected inequalities.}
    Each region $\omega \in \Pi^{\leq l}$ can be defined by linear inequalities 
    in the input space derived from the hyperplanes of every layer. Construct 
    a sign vector from the activation pattern by mapping $(0, 1) \to (1, -1)$:
    \begin{equation}
        \boldsymbol{\sigma}^l_\omega = -2\pi^l_\omega + \mathbf{1},
        \qquad 
        \Sigma^l_\omega = \mathrm{diag}(\boldsymbol{\sigma}^l_\omega).
    \end{equation}
    Under this convention an active neuron ($\pi^l_{\omega,i} = 1$, with
    pre-activation $z_i > 0$) has $\sigma_i = -1$, so the sign-corrected
    pre-activation $\sigma_i z_i \leq 0$. The layer-$l$ constraint defining
    the closure of $\omega$ (the boundaries $z_i = 0$ have measure zero) is
    therefore
    \begin{equation}
        \Sigma^l_\omega \big( W^l A^{l-1}_\omega \mathbf{x} 
            + W^l \mathbf{c}^{l-1}_\omega + \mathbf{b}^l \big) \leq 0.
    \end{equation}
    Aggregating the constraints from all layers up to $l$ fully characterizes 
    $\omega$. Writing 
    $D^l_\omega = \Sigma^l_\omega W^l A^{l-1}_\omega$ and 
    $\mathbf{g}^l_\omega = -\Sigma^l_\omega (W^l \mathbf{c}^{l-1}_\omega + \mathbf{b}^l)$, 
    the region takes the half-space form
    \begin{equation}
        \omega = \{ \mathbf{x} \in \mathcal{X} :
                    D^l_\omega \mathbf{x} \leq \mathbf{g}^l_\omega \}.
    \end{equation}

    \paragraph{Data-supported regions in practice.}
    None of the estimators needs the half-space form or an enumeration of
    $\Pi^{\leq l}$. For each $\mathbf{x} \in \mathcal{D}$ a single forward
    pass records the patterns $\pi^1(\mathbf{x}), \dots, \pi^l(\mathbf{x})$;
    their concatenation is hashed, and points with the same hash lie in the
    same region of $\Pi^{\leq l}$. The distinct hashes are
    $\Omega^{\leq l}_{\mathcal{D}}$, and counting them per label gives the
    table $n_{y,\omega}$ of \cref{eq:plugin}. The pattern of the first point
    seen in a region is stored as its representative and is the input to the
    functional quotient (\cref{app:func_algorithm}). The half-space form
    above describes the geometry of a region but is not used by the
    estimators. In order to find, characterize and use these regions, we have developed the software \texttt{parx}~\cite{parx}.

\newpage

\section{The 2D Composite Dataset}
\label{app:composite_dataset}
 
The Composite dataset is designed to provide a controlled but
heterogeneous testbed for partition-based information analysis. The
two canonical 2D classification benchmarks --- two-moons and
concentric circles --- each test a single topological feature
(curvature in the moons, nested manifolds in the circles); separately
they are too uniform to exercise the routing estimator across regimes
of varying density, class entropy, and local linear separability. We
combine them, together with isotropic Gaussian blobs, into a single
seven-class joint distribution that mixes the three structurally
distinct generators within a shared two-dimensional input domain.
 
\subsection{Generators}
 
The dataset contains $N = 10\,000$ samples across $|\mathcal{Y}| = 7$
classes drawn from three generative components:
 
\paragraph{Moons (classes 0--1, 40\% of samples).}
We sample $N_1 = 4\,000$ points (split equally between the two
classes) from the standard \texttt{make\_moons} generator with noise
$\sigma = 0.10$, producing two interlocking half-moons, translated by
$(+0.5, +0.5)$. The resulting
support is concentrated along two one-dimensional curves with mild
isotropic spread; the two classes have non-linearly entangled support
that requires at least one ReLU bend to separate.
 
\paragraph{Concentric circles (classes 2--3, 40\% of samples).}
We sample $N_2 = 4\,000$ points (split equally) from the standard
\texttt{make\_circles} generator with inner-to-outer scale factor
$0.75$ and noise $\sigma = 0.05$. The raw output, supported on the
unit disk, is then scaled by $3.0$ and translated by $(+1, +1)$, so
that both circles enclose the moons without overlapping them
(\cref{fig:composite_dataset}). The two classes form nested closed
loops; the inner class cannot be linearly separated from the outer,
and neither circle can be separated from the moons by a single
hyperplane.
 
\paragraph{Gaussian blobs (classes 4--6, 20\% of samples).}
We sample $N_3 = 2\,000$ points ($667$, $667$ and $666$ per class) from
three isotropic Gaussians with $\sigma = 0.3$ centred at $(-2, 4)$,
$(3, -3)$, and $(5, 4)$, outside the outer circle. The three classes
are well separated from each other and from the moons and circles.
They serve as an ``easy'' sub-task.

After concatenation and the train/test split (\cref{app:train_test}),
the inputs are min-max scaled component-wise to $[-1, 1]^2$, with the
scaling fitted on the training split, bounding the input domain for
stable neural-network initialisation.
 
\subsection{Class statistics and label entropy}
 
The three generators produce non-uniform class proportions: the four
moons-and-circles classes each carry $p = 1/5$ of the mass, while the
three blob classes each carry $p = 1/15$, summing exactly to $1$. The
resulting class entropy is
\begin{equation}
  H(Y)
  \;=\;
  -\,4 \cdot \tfrac{1}{5} \log_2 \tfrac{1}{5}
  \;-\; 3 \cdot \tfrac{1}{15} \log_2 \tfrac{1}{15}
  \;\approx\; 2.639 \;\text{bits},
\end{equation}
which is the data-side ceiling appearing as a horizontal reference
line on the Composite panel of \cref{fig:layer-profile} in the main
text; the stratified held-out split has the same value,
$\res{hy.composite.heldout}$ bits. Note that
$H(Y) < \log_2 7 \approx 2.807$: the dataset
is deliberately not class-balanced, which gives a non-trivial
upper bound that the routing estimator must respect.
 
\subsection{Properties relevant to partition-based analysis}
 
Three properties of the Composite dataset matter for our experiments:
 
\paragraph{Heterogeneous local geometry.}
The three generators have very different supports in the $[-1, 1]^2$
domain: the blobs concentrate mass in three small isotropic regions
near the corners, the moons stretch along two curves in the centre,
and the circles form two concentric rings around the moons. A trained network must produce
class-discriminative routing across all three regimes, so the
data-supported regions $\Omega_{\mathcal{D}}$ are not all of the same
size or shape. This stresses the routing estimator more than a
single-generator dataset would.
 
\paragraph{Rich class entropy at low ambient dimension.}
With two input dimensions and $N = \res{heldout.composite.N_heldout}$
held-out points, every Composite network stays in the well-sampled
regime (occupancy $\rho < \res{occupancy.composite.rho_max}$ at the
deepest layer, \cref{app:occupancy}), and at the same time the seven
classes give a non-trivial $H(Y) = 2.64$ bits to recover. Composite is
therefore the dataset on which the routing estimator can be followed
through a multi-bit range without approaching saturation.
 
\paragraph{Mixed difficulty per class.}
The blob classes are separated from the rest by a few linear cuts,
while the moons and circles classes require non-linear bends. The
dataset therefore mixes sub-tasks that a narrow network solves early
with sub-tasks that need more layers and more training.

\subsection{Train/test handling}
\label{app:train_test}

Composite and WBC are split once into a training set ($80\%$) and a
held-out test set ($20\%$), stratified by class and fixed across seeds and
architectures; MNIST FC uses the standard MNIST training and test splits
(\cref{app:wbc_dataset,app:mnist_fc_dataset}). All preprocessing (min-max
scaling, and PCA for MNIST) is fitted on the training set and applied
unchanged to the test set, whose values are clipped to $[-1, 1]$. Networks are trained
on the training set only. Every information-theoretic quantity in the paper
--- the counts $n_{y,\omega}$, $n_y$, $n_\omega$ of the routing estimator,
the data support $\Omega_{\mathcal{D}}$ of the functional quotient, and all
three baselines --- is computed on the held-out set, so $N = |\mathcal{D}|$
is $\res{heldout.composite.N_heldout}$ for Composite,
$\res{heldout.wbc.N_heldout}$ for WBC and $10{,}000$ for MNIST FC. The
partition is thus evaluated on points it was not fitted to, at the cost of
smaller count tables, and hence of larger occupancy
$\rho = |\Omega_{\mathcal{D}}|/N$. \Cref{app:heldout} compares this with the
in-sample protocol on all points: at the deepest layer the plug-in estimate
moves by at most $\res{heldout.composite.max_abs_delta_plug_in.deepest}$
bits on Composite and $\res{heldout.wbc.max_abs_delta_plug_in.deepest}$
bits on WBC.

\begin{figure}[h!]
  \centering
  \includegraphics[width=0.75\textwidth]{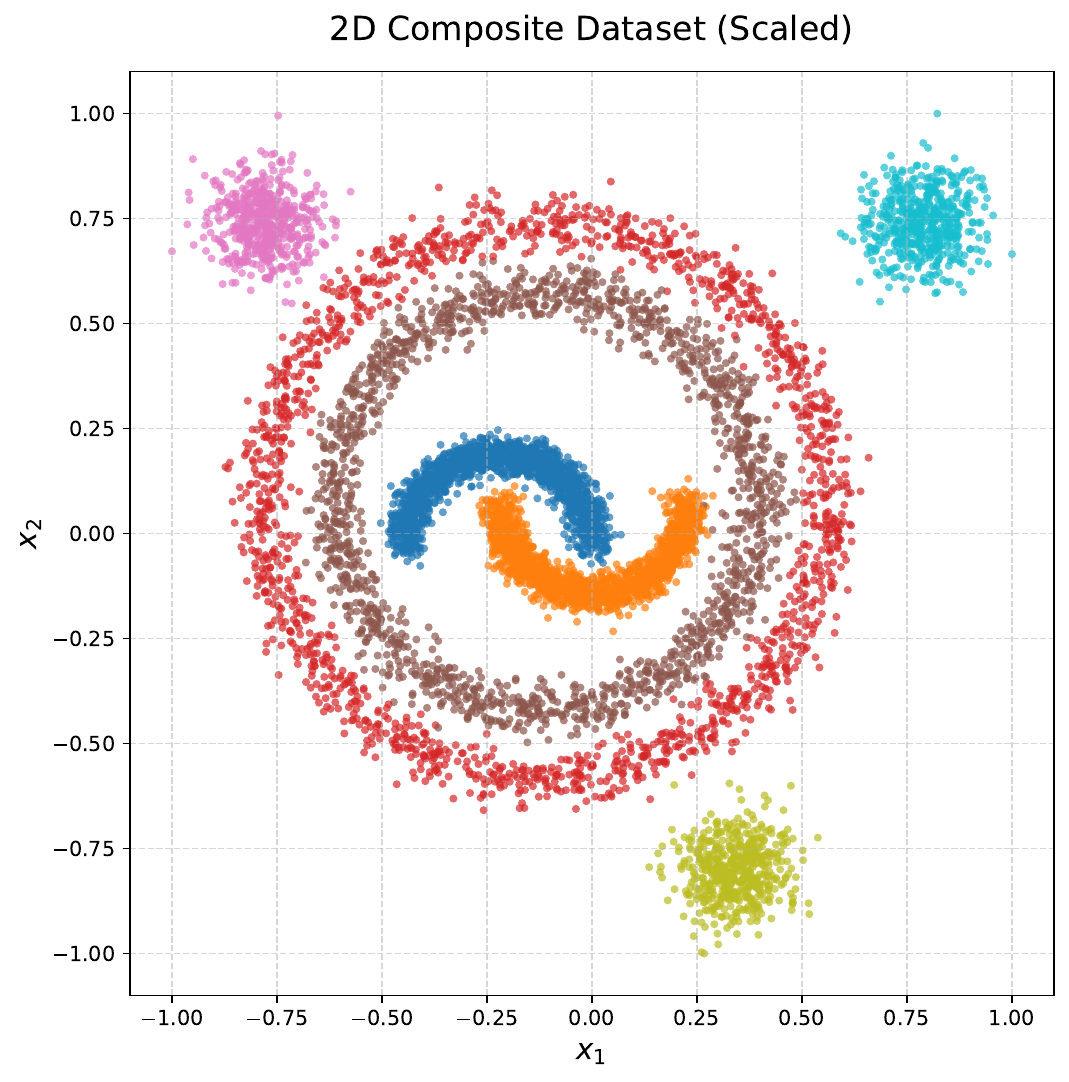}
  \caption{A realisation of the seven-class 2D Composite dataset
    after min-max scaling to $[-1, 1]^2$. Classes 0--1 (moons) and
    2--3 (circles) each carry $20\%$ of the mass; classes 4--6
    (blobs) each carry $\sim 6.7\%$. The two circles enclose the
    moons and the blobs lie outside the circles; no two generators
    overlap. This gives the heterogeneous routing geometry analysed in
    \cref{sec:calibration-scatter,sec:calibration-layerwise}.}
  \label{fig:composite_dataset}
\end{figure}

\Cref{fig:composite_dataset} shows a representative realisation of
the dataset.
 
\section{The WBC Dataset}
\label{app:wbc_dataset}

The Wisconsin Breast Cancer (Diagnostic) dataset
\citep{williamwolbergBreastCancerWisconsin1993} contains $569$ samples
with $30$ real-valued features: the mean, standard error and largest
value of ten characteristics of cell nuclei measured in digitised
images of fine-needle aspirates. The binary label is the diagnosis,
with malignant ($212$ samples) as the positive class and benign ($357$)
as the negative class, so $H(Y) = \res{hy.wbc.insample}$ bits on all
points and $\res{hy.wbc.heldout}$ bits on the
$\res{heldout.wbc.N_heldout}$ held-out points. The features are used
without dimensionality reduction and min-max scaled to $[-1, 1]^{30}$
as described in \cref{app:train_test}. WBC is the smallest dataset in
the study and the only one with a binary label: it tests the routing
estimator at low label entropy and small $N$, where the occupancy
$\rho$ grows quickly with the width (\cref{app:occupancy}). The
architectures and the training protocol are the same as for Composite
(\cref{sec:calibration-setup}).

\section{The MNIST FC Dataset}
\label{app:mnist_fc_dataset}
 
The MNIST FC dataset complements the synthetic Composite and the
small clinical WBC datasets used in our main calibration experiments.
It introduces a third, structurally distinct setting: a $10$-class
image classification task whose label entropy ($H(Y) \approx 3.32$
bits) sits well above the other two datasets, and whose input is a
genuine image rather than a low-dimensional synthetic vector.
 
\subsection{Pipeline}
 
We start from the standard MNIST handwritten-digit dataset
\citep{lecun1999mnist}: $60\,000$ training images and $10\,000$ test
images of $28 \times 28$ greyscale digits, labelled $0$--$9$. We
flatten each image to a vector in $[0, 1]^{784}$ and apply the
following two-step preprocessing:
 
\paragraph{1. PCA to ten dimensions.}
We fit a principal-component analysis on the training set and project
both train and test sets to the top $d_{\mathrm{PCA}} = 10$ principal
components. The choice of $d_{\mathrm{PCA}} = 10$ is deliberate: it
keeps the input dimension low enough that narrow networks stay in the
well-sampled regime ($|\Omega_{\mathcal{D}}| \ll N$) while retaining
about half ($48.8\%$) of the training-set pixel variance. Larger
$d_{\mathrm{PCA}}$ pushes networks of the same width into the
fine-resolution regime ($|\Omega_{\mathcal{D}}| \approx N$) of
\cref{subsec:functional_quotient} where $\hat I$ becomes
uninformative; smaller $d_{\mathrm{PCA}}$ discards class-relevant
directions, so the input itself carries less information about the
label (test accuracy falls below $50\%$ at $d_{\mathrm{PCA}} = 2$,
\cref{fig:mnist_metrics}). The study of \cref{sec:capacity-setup}
varies $d_{\mathrm{PCA}}$ from $2$ to $20$ with the same pipeline.
 
\paragraph{2. Min-max scaling to $[-1, 1]^{10}$.}
The PCA-projected coordinates are min-max scaled component-wise so
that each of the ten dimensions lies in $[-1, 1]$. This matches the
Composite preprocessing and provides comparable input scales across
datasets.
 
The PCA basis and min-max bounds are fitted on the training set and
applied unchanged to the test set, to avoid information leakage. All
information-theoretic quantities reported in the main paper are
computed on the $10\,000$-sample test set, on which
$H(Y) = \res{hy.mnist.heldout}$ bits (the test split is approximately
balanced across digits but not exactly so).
 
\subsection{Architectures}
 
We use three narrow fully-connected ReLU architectures of three
hidden layers each: $[3, 3, 3]$, $[5, 5, 5]$, and $[7, 7, 7]$. The
narrow widths are dictated by the requirement
$|\Omega_{\mathcal{D}}| \ll N$ at the deepest layer: with
$d_{\mathrm{PCA}} = 10$ the occupancy at the last hidden layer is
$\rho = 0.122$ at width $7$, but $0.743$ at width $15$ and $0.954$ at
width $25$ (\cref{tab:occupancy_mnist}), where most test samples occupy
their own region and the routing estimator becomes uninformative.
Restricting to the well-sampled regime is therefore not a cherry-pick
but a stated scope of the framework; the wider networks are the
subject of the saturation study (\cref{sec:capacity-results}).
 
The $[3, 3, 3]$ architecture is included in the calibration scatter
of \cref{fig:calibration-scatter} (where it provides additional data
points across the lower bit range), but its partition is very coarse
($\rho = 0.004$, about $40$ regions for $10\,000$ points at the deepest
layer) and the network reaches only about $60\%$ test accuracy
(\cref{fig:mnist_metrics}); we therefore exclude it from the layerwise
profile of \cref{fig:layer-profile}. The saturation study of
\cref{sec:capacity-results} trains $[7,7,7]$ and the wider
$[15,15,15]$, $[25,25,25]$ and $[50,50,50]$ at every PCA dimension
$d \in \{2, 3, 4, 5, 10, 15, 20\}$ (\cref{sec:capacity-setup}); its
results are reported separately in
\cref{fig:mnist-capacity,fig:mnist-rho-eps}.
 
Training follows the same SGD protocol as Composite/WBC ($150$ epochs,
$\alpha = 10^{-3}$, momentum $0.9$, batch size $32$, cross-entropy,
no weight decay or dropout). We use five seeds per architecture,
giving $15$ networks total.

\newpage

\section{Probabilistic Details for Section~\ref{sec:framework}}
\label{app:joint}

    \subsection{Joint factorization and pushforwards}

        Under the DAG $Y \to X \to (\Pi, T)$ with $\Pi, T$ deterministic 
        functions of $X$ via $\Pi = \pi(X)$ and 
        $T = \phi_\Pi(X) = A_\Pi X + \mathbf{c}_\Pi$, the joint law factorizes as
        \begin{equation}
            p_{YX\Pi T}(y, x, \omega, t) 
                = p_Y(y)\, p_{X \mid Y}(x \mid y)\, 
                  \delta_{\omega, \pi(x)}\, \delta(t - \phi_\omega(x)).
        \end{equation}
        The induced marginals are
        \begin{equation}
            p_\Pi(\omega) = \int_\omega p_X(x)\, dx,
            \qquad 
            p_{Y \mid \Pi}(y \mid \omega) 
                = \frac{\int_\omega p_{XY}(x, y)\, dx}{p_\Pi(\omega)},
        \end{equation}
        and the within-region activation law is the pushforward of the 
        within-region input law through the region's affine map,
        \begin{equation}
            \nu_\omega := p_{T \mid \Pi}(\,\cdot\, \mid \omega) 
                = (\phi_\omega)_\# p_{X \mid \Pi}(\,\cdot\, \mid \omega).
            \label{eq:pushforward}
        \end{equation}
        Concretely, for any Borel $E \subseteq \mathbb{R}^{n_l}$, 
        $\Pr(T \in E \mid \Pi = \omega) = \Pr(X \in \phi_\omega^{-1}(E) \mid \Pi = \omega)$. 
        The marginal of $T$ is the mixture
        \begin{equation}
            p_T = \sum_{\omega \in \Pi^{\leq l}} p_\Pi(\omega)\, \nu_\omega.
            \label{eq:T_mixture}
        \end{equation}

        \paragraph{Remark on rank.}
        The pushforward $\nu_\omega$ is always well-defined as a measure, 
        regardless of $\mathrm{rank}(A_\omega)$. If $A_\omega$ has rank $r < n_l$, 
        then $\nu_\omega$ is supported on an $r$-dimensional affine subspace 
        of $\mathbb{R}^{n_l}$ and is singular with respect to Lebesgue measure; 
        if $A_\omega = 0$, then $\nu_\omega = \delta_{\mathbf{c}_\omega}$. All 
        MI terms involving $\nu_\omega$ remain finite and well-defined in every 
        case because $Y$ takes finitely many values; the pathologies of 
        differential entropy that afflict $I(Y; T)$ in general IB analyses 
        do not apply to the decomposition terms of 
        Theorem~\ref{thm:cpwl_decomp}.

    \subsection{Per-region form of the decomposition terms}
    \label{app:per_region}

        Because $\Pi$ measurably partitions $\mathcal{X}$, three of the four 
        terms in Theorem~\ref{thm:cpwl_decomp} localize per-region. Let 
        $P_\omega$ denote orthogonal projection onto $\mathrm{row}(A_\omega)$ 
        and $P_\omega^\perp$ onto $\ker A_\omega$. Within region $\omega$, 
        $T = A_\omega X + \mathbf{c}_\omega$ depends on $X$ only through 
        $P_\omega X$, so conditioning on $T$ inside region $\omega$ is 
        equivalent (up to the affine bijection $A_\omega|_{\mathrm{row}(A_\omega)}$) 
        to conditioning on $P_\omega X$.

        \begin{align}
            I(Y; \Pi) 
                &= \sum_\omega p_\Pi(\omega) 
                   \sum_y p_{Y \mid \Pi}(y \mid \omega) 
                   \log \frac{p_{Y \mid \Pi}(y \mid \omega)}{p_Y(y)}, \\
            I(Y; T \mid \Pi) 
                &= \sum_\omega p_\Pi(\omega)\, I_\omega(Y; A_\omega X) 
                 = \sum_\omega p_\Pi(\omega)\, I_\omega(Y; P_\omega X), \\
            I(Y; X \mid \Pi, T) 
                &= \sum_\omega p_\Pi(\omega)\, 
                   I_\omega(Y; P_\omega^\perp X \mid P_\omega X),
            \label{eq:repr_insuff}
        \end{align}
        with $I_\omega$ denoting MI under the within-region law 
        $p_{\cdot \mid \Pi = \omega}$. The conditioning on $P_\omega X$ in 
        the last line is essential: a region $\omega$ is a bounded convex 
        polytope, so the row-space and kernel components $P_\omega X$ and 
        $P_\omega^\perp X$ are in general statistically dependent under 
        $p_{X \mid \Pi = \omega}$, and 
        $I_\omega(Y; P_\omega^\perp X \mid P_\omega X) \neq I_\omega(Y; P_\omega^\perp X)$. 
        The representation insufficiency $I(Y; X \mid \Pi, T)$ vanishes iff 
        $Y \perp P_\omega^\perp X \mid P_\omega X$ under 
        $p_{X \mid \Pi = \omega}$ for every region $\omega$.

        The routing loss does not localize per-region because it couples 
        regions via pushforward overlap:
        \begin{equation}
            I(Y; \Pi \mid T) 
                = \int p_T(t) \sum_y p_{Y \mid T}(y \mid t) 
                  \sum_{\omega \in S(t)} 
                    p_{\Pi \mid Y, T}(\omega \mid y, t) 
                    \log \frac{p_{\Pi \mid Y, T}(\omega \mid y, t)}
                              {p_{\Pi \mid T}(\omega \mid t)}\, dt,
        \end{equation}
        where $S(t) = \{\omega : t \in \mathrm{supp}(\nu_\omega)\}$.

    \subsection{Proof of Theorem~\ref{thm:cpwl_decomp}}
    \label{app:proof_cpwl_decomp}

        Applying the chain rule to $I(Y; \Pi, T)$ in two orders,
        \begin{align}
            I(Y; \Pi, T) &= I(Y; \Pi) + I(Y; T \mid \Pi), \\
            I(Y; \Pi, T) &= I(Y; T) + I(Y; \Pi \mid T).
        \end{align}
        Equating and solving for $I(Y; T)$ gives~\eqref{eq:main_decomp}.

        For~\eqref{eq:joint_decomp}: since $H(\Pi, T \mid X) = 0$, chain rule 
        gives $I(Y; X, \Pi, T) = I(Y; X)$ and 
        $I(Y; X, \Pi, T) = I(Y; \Pi, T) + I(Y; X \mid \Pi, T)$, whence 
        $I(Y; \Pi, T) = I(Y; X) - I(Y; X \mid \Pi, T)$. \qed

    \subsection{Proofs of Corollaries~\ref{cor:upper_bound} and~\ref{cor:balance}}
    \label{app:proof_corollaries}

        \paragraph{\Cref{cor:upper_bound}.} By the chain rule,
        $I(Y; \Pi, T) = I(Y; T) + I(Y; \Pi \mid T) \geq I(Y; T)$, with
        equality iff $I(Y; \Pi \mid T) = 0$. By~\eqref{eq:joint_decomp},
        $I(Y; \Pi, T) = I(Y; X) - I(Y; X \mid \Pi, T) \leq I(Y; X)$, with
        equality iff $I(Y; X \mid \Pi, T) = 0$, i.e.\ iff
        $Y \perp X \mid (\Pi, T)$, which is sufficiency of $(\Pi, T)$
        for $Y$. \qed

        \paragraph{\Cref{cor:balance}.} Subtracting $I(Y; \Pi)$ from both
        sides of~\eqref{eq:main_decomp} gives
        $I(Y; T) - I(Y; \Pi) = I(Y; T \mid \Pi) - I(Y; \Pi \mid T)$, so
        $I(Y; T) > I(Y; \Pi)$ iff
        $I(Y; T \mid \Pi) > I(Y; \Pi \mid T)$. \qed

    \subsection{Proof of Proposition~\ref{prop:folding_loss}}
    \label{app:proof_folding_loss}

        \paragraph{Setup.}
        From~\eqref{eq:T_mixture}, the posterior over regions given $T = t$ is
        \begin{equation}
            p_{\Pi \mid T}(\omega \mid t) 
                = \frac{p_\Pi(\omega)\, \nu_\omega(t)}
                       {\sum_{\omega'} p_\Pi(\omega')\, \nu_{\omega'}(t)},
        \end{equation}
        where densities $\nu_\omega(t)$ are taken with respect to $p_T$ 
        as the dominating measure. Define 
        $S(t) = \{\omega : t \in \mathrm{supp}(\nu_\omega)\}$ and 
        $O = \{t : |S(t)| \geq 2\}$; the posterior is non-degenerate exactly on 
        $O$.

        \paragraph{KL form.}
        The standard expansion of conditional mutual information gives
        \begin{equation}
            I(Y; \Pi \mid T) 
                = \mathbb{E}_T \!\left[ 
                    \sum_{\omega \in S(T)} p_{\Pi \mid T}(\omega \mid T)\, 
                    D_{\mathrm{KL}}\!\big(p_{Y \mid \Pi = \omega, T} 
                                   \,\|\, p_{Y \mid T}(\cdot \mid T)\big) 
                  \right],
            \label{eq:KL_form}
        \end{equation}
        where $p_{Y \mid \Pi = \omega, T}(\cdot \mid t)$ is the joint 
        region-and-activation conditional law of $Y$, and 
        $p_{Y \mid T}(\cdot \mid t) = \sum_{\omega'' \in S(t)} 
        p_{\Pi \mid T}(\omega'' \mid t)\, p_{Y \mid \Pi = \omega'', T}(\cdot \mid t)$ 
        is its mixture over regions in the active set.

        \paragraph{($\Leftarrow$).}
        Suppose distinct $\omega, \omega'$ have 
        $\mathrm{supp}(\nu_\omega) \cap \mathrm{supp}(\nu_{\omega'})$ of positive 
        $p_T$-mass and 
        $p_{Y \mid \Pi = \omega, T = t} \neq p_{Y \mid \Pi = \omega', T = t}$ on 
        a positive-mass subset $B \subseteq O$. On $B$ the mixture 
        $p_{Y \mid T}(\cdot \mid t)$ cannot equal both 
        $p_{Y \mid \Pi = \omega, T = t}$ and 
        $p_{Y \mid \Pi = \omega', T = t}$ since these differ; hence at least 
        one KL term in~\eqref{eq:KL_form} is positive on $B$, and 
        $I(Y; \Pi \mid T) > 0$.

        \paragraph{($\Rightarrow$).}
        If $I(Y; \Pi \mid T) > 0$, then by~\eqref{eq:KL_form} there is a set 
        $B$ with $p_T(B) > 0$ on which the inner sum is strictly positive. 
        For each $t \in B$: (i) positivity requires $|S(t)| \geq 2$, so 
        $B \subseteq O$; (ii) positivity requires at least two 
        $\omega, \omega' \in S(t)$ with 
        $p_{Y \mid \Pi = \omega, T = t} \neq p_{Y \mid \Pi = \omega', T = t}$ 
        (else every conditional equals the mixture and all KLs vanish). 
        Both conditions hold on a positive-mass set. \qed

        \paragraph{Marginal sufficient condition.}
        The proposition is stated in terms of the conditional laws 
        $p_{Y \mid \Pi = \omega, T = t}$ at the overlap. A simpler 
        sufficient condition is that the overlapping regions have differing 
        \emph{marginal} region-conditional label laws 
        $p_{Y \mid \Pi = \omega} \neq p_{Y \mid \Pi = \omega'}$, provided the 
        within-region label distribution is independent of $T$, i.e.\ 
        $Y \perp T \mid \Pi$. Under that independence the two formulations 
        coincide, and the marginal version recovers the older statement that 
        ``two overlapping regions with differing class content contribute to 
        routing loss''. In the general case the conditional version is 
        required: two regions can have identical marginal label laws but 
        differ at a specific overlapping coordinate $t$, which is enough 
        for the routing loss to be positive there.

        \paragraph{Geometric reading.} 
        The proposition factors label-relevant pushforward overlap into two 
        independent ingredients: 
        \emph{image overlap} ($p_T(O) > 0$) -- two regions with distinct 
        affine maps produce overlapping activation-space images -- and 
        \emph{label-content disagreement at the overlap} -- the conditional 
        label distributions at overlapping points $t$ differ between regions. 
        Routing loss is positive iff both hold simultaneously. Overlap 
        between regions whose conditional class content at the overlap 
        agrees (e.g.\ regions carrying the same class from disconnected 
        parts of the input distribution) contributes zero to 
        $I(Y; \Pi \mid T)$: such overlap preserves the conditional label 
        distribution at $t$ and is invisible to the routing loss, even 
        though it is visible to geometric measures of partition 
        complexity~\citep{lewandowskiExploitingSpaceFolding2026}.

    \subsection{Truncation gap of the plug-in routing estimator}
    \label{app:proof_routing_est}

        The plug-in estimator~\eqref{eq:plugin} is the standard empirical MI, 
        obtained by substituting maximum-likelihood counts into the Shannon 
        formula. Its target depends on whether one views the count tables as 
        a sample from $p_{Y, \Pi}$ (giving $I(Y; \Pi)$ in the limit) or as 
        the distribution of $(Y, \Pi)$ restricted to the observed support 
        $\Omega_{\mathcal{D}}$ (giving $I(Y; \Pi \mid \Pi \in \Omega_{\mathcal{D}})$). 
        The two limits differ by the truncation gap analysed below.

        \paragraph{Truncation identity.} Let 
        $U = \{\Pi \notin \Omega_{\mathcal{D}}\}$ be the event that a fresh 
        sample hits an unobserved region, and let 
        $\mathbf{1}_U \in \{0, 1\}$ be its indicator. The classical 
        truncation identity gives
        \begin{equation}
            I(Y; \Pi) 
                \;=\; I(Y; \mathbf{1}_U) 
                + \Pr(U^c)\, I(Y; \Pi \mid U^c) 
                + \Pr(U)\, I(Y; \Pi \mid U).
            \label{eq:truncation_id}
        \end{equation}
        The plug-in estimator $\hat I(Y; \Omega_{\mathcal{D}})$, computed 
        from empirical frequencies that sum to one over $\Omega_{\mathcal{D}}$, 
        targets the unscaled term $I(Y; \Pi \mid U^c)$ (the population MI 
        \emph{conditional on} $\Pi$ being observed) rather than the scaled 
        term $\Pr(U^c)\, I(Y; \Pi \mid U^c)$ that appears 
        in~\eqref{eq:truncation_id}. Hence $\hat I$ is in general 
        \emph{not} a population lower bound for $I(Y; \Pi)$: because 
        conditioning on a subset can change MI in either direction, 
        $I(Y; \Pi \mid U^c)$ can lie above or below $I(Y; \Pi)$ in 
        different regimes.

        \paragraph{Gap bound.} For the third term in~\eqref{eq:truncation_id}, 
        $0 \leq I(Y; \Pi \mid U) \leq H(Y \mid U) \leq H(Y)$, so 
        $\Pr(U)\, I(Y; \Pi \mid U) \leq H(Y)\, \Pr(U)$. For the first term, 
        $I(Y; \mathbf{1}_U) \leq H(\mathbf{1}_U) = h_2(\Pr(U))$ where $h_2$ 
        is the binary entropy, which scales as $-\Pr(U) \log_2 \Pr(U)$ as 
        $\Pr(U) \to 0$. Together,
        \begin{equation}
            \big| I(Y; \Pi) - \Pr(U^c)\, I(Y; \Pi \mid U^c) \big|
                \;\leq\; H(Y)\, \Pr(U) + h_2(\Pr(U)).
            \label{eq:trunc_bound}
        \end{equation}
        In the well-sampled regime $\Pr(U) \to 0$ both terms vanish and 
        $\hat I(Y; \Omega_{\mathcal{D}}) \to I(Y; \Pi \mid U^c) \approx 
        I(Y; \Pi)$. The chain~\eqref{eq:chain} therefore holds up to 
        \eqref{eq:trunc_bound}; we use ``$\approx$'' in~\eqref{eq:chain} 
        to mark this approximate (rather than provable) lower-bound 
        relationship. The right-hand inequalities
$I(Y; \Pi, T) \leq I(Y; X) \leq H(Y)$ are exact: the first is
\cref{cor:upper_bound}, the second holds because $Y$ is discrete.

    \subsection{Proof of Proposition~\ref{prop:func_quotient}}
    \label{app:proof_func_quotient}

        Let $q : \Pi^{l} \to \Pi_{\mathrm{func}}$ be the quotient map. 
        Since $q$ is deterministic, $Y \to \Pi \to \Pi_{\mathrm{func}}$ is 
        Markov and the data-processing inequality gives 
        $I(Y; \Pi_{\mathrm{func}}) \leq I(Y; \Pi)$.

        For equality: 
        $I(Y; \Pi) = I(Y; \Pi_{\mathrm{func}}) + I(Y; \Pi \mid \Pi_{\mathrm{func}})$, 
        so the two coincide iff $I(Y; \Pi \mid \Pi_{\mathrm{func}}) = 0$, i.e., 
        iff $Y \perp \Pi \mid \Pi_{\mathrm{func}}$. This holds iff 
        $p_{Y \mid \Pi = \omega}$ is constant on each equivalence class of 
        $\sim_{\mathrm{func}}$, i.e., iff $p_{Y \mid \Pi = \omega}$ depends on 
        $\omega$ only through $A^l_\omega$.

        \paragraph{Remark.}
        The condition asks $p_{Y \mid \Pi = \omega}$ to be constant on each 
        functional-equivalence class. Two cases are worth distinguishing.

        \emph{(i) Predicted label.} If $Y$ in the proposition is taken to be
the network's own predicted label $\widehat Y = \arg\max \mathcal{N}(X)$,
the condition is still not automatic. Two functionally-equivalent
regions share the linear part $A^l_\omega$ but in general not the
offset $\mathbf{c}^l_\omega$, and they receive different inputs, so
their images $\phi^l_\omega(\omega)$ differ and the head can assign
them different predicted labels. The condition holds for the
predicted label when, in addition, equivalent regions have equal
offsets and the head is constant on the union of their images.

        \emph{(ii) Ground-truth label.} If $Y$ is the data label, two 
        functionally-equivalent regions can still differ: the regions can lie 
        in disjoint parts of the input space and carry different ground-truth 
        class proportions in $p_{X, Y \mid \Pi = \omega}$ even though the 
        linear operator $A^l_\omega$ is the same. The equality condition then 
        additionally requires $p_{Y \mid X, \Pi = \omega}$ to depend on $X$ 
        only through $A^l_\omega X$ -- i.e.\ no $Y$-relevant variation along 
        $\ker A^l_\omega$ -- a property of the data distribution, not of the 
        network alone. All reported estimates use the ground-truth label,
for which the quotient is a lower bound but equality is not
guaranteed. The gap
$I(Y; \Pi) - I(Y; \Pi_{\mathrm{func}}) = I(Y; \Pi \mid \Pi_{\mathrm{func}})$
is what the difference $\hat I - \hat I_{\mathrm{func}}$ estimates, and
it is reported with every quotient estimate
(\cref{fig:mnist-capacity,tab:label_permutation}).
\qed

    \subsection{Relation to prior IB literature}
    \label{app:prior_ib}

        The pathology of $I(Y; T)$ for deterministic continuous networks has 
        been discussed by \citet{saxeInformationBottleneckTheory2019} and
\citet{goldfeldEstimatingInformationFlow2019}, who show that estimates of
it are dominated by binning choices; \citet{goldfeldEstimatingInformationFlow2019}
inject noise into the network to make the quantity well-posed. Our approach is 
        orthogonal: we analyze the joint $(\Pi, T)$ whose combinatorial 
        component is entropy-regular by construction, giving finite exact 
        identities without altering the model. The price is tracking the 
        routing-loss term $I(Y; \Pi \mid T)$, whose geometric interpretation 
        via Proposition~\ref{prop:folding_loss} is of independent interest.

\newpage

\section{Clustering Regions by Functional Equivalence}
\label{app:func_algorithm}

    Given $\Omega_{\mathcal{D}}$, network weights, and tolerance 
    $\varepsilon \in [0, 2]$, we group regions by the relative Frobenius 
    criterion of \eqref{eq:func_equiv}: 
    $\omega \sim_\varepsilon \omega'$ iff $S^\omega_l = S^{\omega'}_l$ and 
    $\|\tilde A^l_\omega - \tilde A^l_{\omega'}\|_F / 
     \tfrac12(\|\tilde A^l_\omega\|_F + \|\tilde A^l_{\omega'}\|_F) 
    \leq \varepsilon$. The active-set equality
$S^\omega_l = S^{\omega'}_l$ is required for the matrix difference to
be well-defined ($\tilde A^l_\omega, \tilde A^l_{\omega'}$ otherwise
have incompatible row counts), so regions are first grouped by
$S^\omega_l$ and compared only within a group. Within a group, each
quotient class is represented by its first member, and a new region
joins the first class whose representative is within $\varepsilon$ of
it.

    For $\varepsilon = 0$ this is exact functional equivalence (raw 
    partition); larger $\varepsilon$ merges progressively more pairs. 
    Because the relation $\sim_\varepsilon$ is not transitive when 
    $\varepsilon > 0$ (regions $\omega_1, \omega_3$ may be more than 
    $\varepsilon$ apart even if 
    $\omega_1 \sim_\varepsilon \omega_2 \sim_\varepsilon \omega_3$), the 
    greedy clustering in \cref{alg:func_cluster} can produce slightly 
    different partitions depending on the order in which regions are 
    visited. We iterate over $\Omega_{\mathcal{D}}$ in first-encounter 
    order, i.e.\ the order in which the regions are first reached by the 
    data points, which is deterministic and reproducible. 
    \Cref{app:ordering} measures the effect of the order over 
    $\res{ordering.n_random_orders}$ random visiting orders: at 
    $\varepsilon = 0.3$ the quotient size varies by a median of 
    $\res{ordering.eps0.3.all_layers.spread_median}$ classes and at most 
    $\res{ordering.eps0.3.all_layers.relative_spread_max.pct}\%$ of 
    $|\Omega_{\mathcal{D}}|$, and the Miller--Madow-corrected quotient MI by 
    at most $\res{ordering.eps0.3.all_layers.miller_madow_func_spread_max}$ bits.

    \begin{algorithm}[H]
        \caption{Functional-quotient clustering}
        \label{alg:func_cluster}
        \begin{algorithmic}[1]
            \Require $\Omega_{\mathcal{D}}$, weights $\{W^i, \mathbf{b}^i\}_{i=1}^l$, 
                     tolerance $\varepsilon \in [0, 2]$
            \Ensure Quotient $\Omega_{\mathcal{D}, \mathrm{func}}$
            \State $\Omega_{\mathcal{D}, \mathrm{func}} \gets \varnothing$
            \For{each $\omega \in \Omega_{\mathcal{D}}$ in first-encounter order}
                \State Read $S^\omega_i$ from $\pi^i_\omega$ for $i = 1, \dots, l$
                \State $\tilde A^l_\omega \gets \prod_{i=l}^{1} W^i[S^\omega_i, S^\omega_{i-1}]$
                \State Search the classes of $\Omega_{\mathcal{D}, \mathrm{func}}$
with active set $S^\omega_l$, in creation order, for the first
class $[\omega']$ whose representative $\omega'$ satisfies
$\omega \sim_\varepsilon \omega'$
\If{found}
\State Add $\omega$ to $[\omega']$
                \Else
                    \State Create new class $\{\omega\}$ in
$\Omega_{\mathcal{D}, \mathrm{func}}$ with representative $\omega$
                \EndIf
            \EndFor
            \State \Return $\Omega_{\mathcal{D}, \mathrm{func}}$
        \end{algorithmic}
    \end{algorithm}

    \paragraph{Complexity.} Each comparison $\omega \sim_\varepsilon \omega'$
costs $O(|S^\omega_l| \cdot n_0)$ for the Frobenius distance. Each region
is compared only with the representatives of existing classes in its
active-set group, so the number of comparisons is at most
$|\Omega_{\mathcal D}| \cdot |\Omega_{\mathcal D, \mathrm{func}}|$, with an
$O(|\Omega_{\mathcal D}|^2)$ worst case at small $\varepsilon$. In our 
    experiments $|\Omega_{\mathcal{D}}|$ reaches 
    $\res{ordering.eps0.3.all_layers.regions_max}$ (MNIST, width $50$); for 
    large supports the comparison 
    can be accelerated by locality-sensitive hashing on rounded 
    $\tilde A^l_\omega$, reducing the cost to near-linear in 
    $|\Omega_{\mathcal{D}}|$.

    \paragraph{Operating range.} The relative criterion is bounded in
$[0, 2]$ and admits a clean interpretation: $\varepsilon = 0$ recovers
the raw partition; $\varepsilon \in [0.2, 0.5]$ is the window in which
$\rho_{\mathrm{func}}$ drops and the quotient pulls a saturated estimate
back below $H(Y)$, provided the network has redundant linear operators
to collapse (\cref{sec:capacity-results} and
\cref{fig:mnist-capacity,fig:mnist-rho-eps}); and $\varepsilon \geq 1$
over-merges. Merging already starts below the window, and the estimate
still moves with $\varepsilon$ inside it (\cref{sec:rho-discussion}),
so $\varepsilon$ is reported with every quotient estimate. Where a single value is reported
(\cref{tab:label_permutation}), we use $\varepsilon = 0.3$
as a representative operating point.

\begin{figure}[h]
      \centering
      \includegraphics[width=\linewidth]{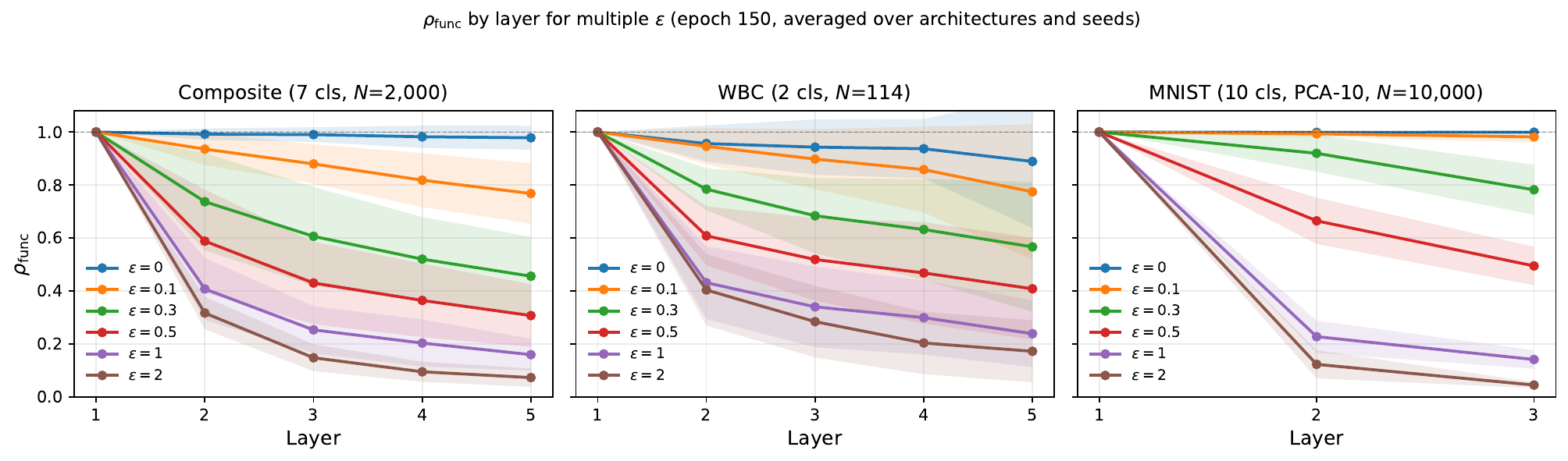}
      \caption{$\rho_{\mathrm{func}}^{\,l}(\varepsilon)$ as a function of hidden-layer depth on the calibration networks (the five-layer architectures of Composite and WBC, and $[5,5,5]$ and $[7,7,7]$ on MNIST FC), one curve per $\varepsilon$, mean $\pm 1\sigma$ across architectures and seeds. $\rho_{\mathrm{func}}^{\,1} = 1$ structurally on every panel; for $\varepsilon > 0$ the mean ratio decreases with depth.}
      \label{fig:rho-func-layerwise}
    \end{figure}

    \paragraph{Layerwise compression ratio.}
    \label{app:func_layerwise}
    \Cref{fig:rho-func-layerwise} reports $\rho_{\mathrm{func}}^{\,l}(\varepsilon)$ 
    as a function of layer depth on the calibration networks. 
    $\rho_{\mathrm{func}}^{\,1} = 1$ for every dataset, architecture and 
    $\varepsilon$: the first hidden layer realises every distinct activation 
    pattern as a distinct linear operator, so functional equivalence is a 
    phenomenon of compositions. For $\varepsilon > 0$, $\rho_{\mathrm{func}}^{\,l}$ 
    then decreases with depth on average --- deeper compositions admit more 
    activation-pattern combinations whose products of selection matrices 
    coincide --- at a rate set by $\varepsilon$. The decrease is not monotone 
    for every network: of the $\res{rho_func_layerwise.n_curves}$ per-network curves 
    ($\res{rho_func_layerwise.n_networks}$ networks, $14$ values of $\varepsilon > 0$), 
    $\res{rho_func_layerwise.n_curves_rising}$ rise somewhere with depth, by at most 
    $\res{rho_func_layerwise.max_rise}$.

\newpage
\section{Finite-Sample Validation of the Routing Estimator}
\label{app:estimator_validation}

This appendix collects the evidence on when the plug-in routing estimator
$\hat I(Y;\Omega_{\mathcal{D}})$ of \cref{eq:plugin} can be trusted. All
numbers are computed on the held-out points (\cref{app:composite_dataset}),
at the last epoch, and averaged over the five seeds; the tables are generated
from the stored results of the released code. The organising quantity is the
occupancy
\begin{equation*}
    \rho := |\Omega_{\mathcal{D}}| / N \in (0, 1],
\end{equation*}
the number of data-supported regions per sample. At small $\rho$ most regions
hold several points and the counts $n_{y,\omega}$ are informative; as
$\rho \to 1$ almost every region holds a single point, the partition becomes an
injective map from samples to regions, and $\hat I$ is pushed towards $H(Y)$
whatever the network has learned (\cref{app:proof_routing_est}).

\subsection{Occupancy of the evaluated networks}
\label{app:occupancy}

\begin{figure}[h]
    \centering
    \includegraphics[width=\linewidth]{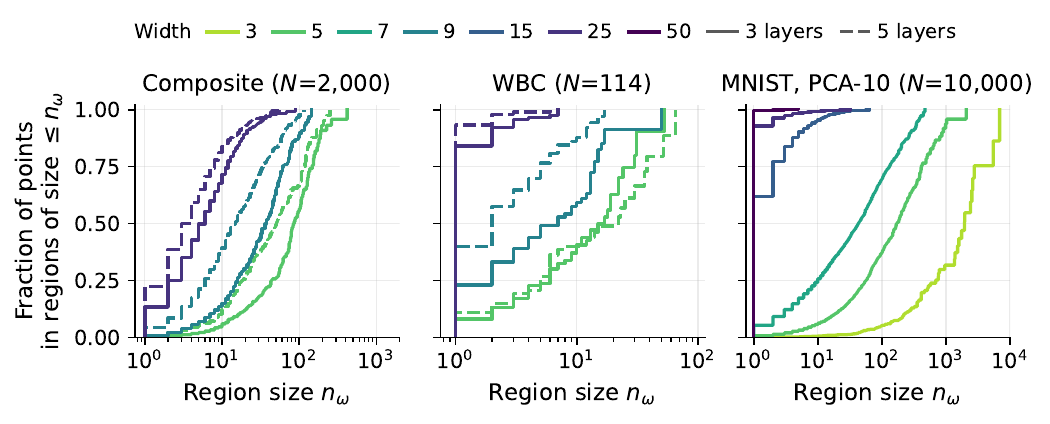}
    \caption{Distribution of region sizes at the deepest layer: the fraction
    of points that lie in a region holding at most $n_\omega$ points. Colour
    is the hidden width, line style the depth; MNIST at PCA dimension $10$.
    Held-out points, last epoch, five seeds pooled. The value at
    $n_\omega = 1$ is the singleton-sample fraction of
    \cref{tab:occupancy}; curves that reach $1$ at small $n_\omega$ belong to
    saturated partitions ($\rho \to 1$).}
    \label{fig:occupancy}
\end{figure}

\Cref{tab:occupancy,tab:occupancy_mnist} report $\rho$ at the deepest layer
for every configuration, together with the fraction of regions that hold a
single point and the fraction of points that sit alone in their region. The
Composite networks stay below $\rho = \res{occupancy.composite.rho_max}$. On
WBC, with only $N = \res{heldout.wbc.N_heldout}$ held-out points, the two
width-$25$ architectures are close to saturation: $\rho =
\res{wbc_rho.[25, 25, 25].heldout}$ for $[25,25,25]$ and $\rho =
\res{wbc_rho.[25, 25, 25, 25, 25].heldout}$ for $[25,25,25,25,25]$ ($\res{wbc_rho.[25, 25, 25].insample}$
and $\res{wbc_rho.[25, 25, 25, 25, 25].insample}$ on all $569$ points). Their calibration
points in \cref{fig:calibration-scatter,fig:layer-profile} lie in the regime
where $\hat I$ is not trustworthy and should be read with that caveat. On
MNIST, $\rho$ grows quickly with the PCA dimension and the width
(\cref{tab:occupancy_mnist}); this is the saturation that the functional
quotient addresses in \cref{sec:capacity-results}. \Cref{fig:occupancy} shows
the full distribution of region sizes behind these numbers. At low $\rho$ most
points sit in regions holding tens of points or more; in the saturated
networks (WBC width $25$, MNIST widths $25$ and $50$) almost every point sits
in a region of size one or two.

\begin{table}[h]
    \centering
    \small
    \caption{Occupancy at the deepest layer, Composite ($N = \res{heldout.composite.N_heldout}$)
    and WBC ($N = \res{heldout.wbc.N_heldout}$), held-out points, mean over seeds. Regions: $|\Omega_{\mathcal{D}}|$; singleton regions: fraction
    of regions holding one point; singleton samples: fraction of points alone in
    their region.}
    \label{tab:occupancy}
\begin{tabular}{llrrrr}
\toprule
Dataset & Architecture & Regions & $\rho$ & Singleton regions & Singleton samples \\
\midrule
Composite & $[5,5,5]$ & 59.2 & 0.030 & 0.123 & 0.004 \\
 & $[5,5,5,5,5]$ & 108.2 & 0.054 & 0.138 & 0.009 \\
 & $[9,9,9]$ & 128.4 & 0.064 & 0.133 & 0.009 \\
 & $[9,9,9,9,9]$ & 294.2 & 0.147 & 0.291 & 0.044 \\
 & $[25,25,25]$ & 607.4 & 0.304 & 0.438 & 0.133 \\
 & $[25,25,25,25,25]$ & 822.2 & 0.411 & 0.540 & 0.222 \\
\midrule
WBC & $[5,5,5]$ & 21.0 & 0.184 & 0.438 & 0.081 \\
 & $[5,5,5,5,5]$ & 23.0 & 0.202 & 0.530 & 0.111 \\
 & $[9,9,9]$ & 41.0 & 0.360 & 0.627 & 0.230 \\
 & $[9,9,9,9,9]$ & 64.0 & 0.561 & 0.705 & 0.398 \\
 & $[25,25,25]$ & 102.6 & 0.900 & 0.931 & 0.840 \\
 & $[25,25,25,25,25]$ & 109.6 & 0.961 & 0.970 & 0.933 \\
\bottomrule
\end{tabular}

\end{table}

\begin{table}[h]
    \centering
    \small
    \caption{Occupancy $\rho$ at the last hidden layer on MNIST
    ($N = 10\,000$), per architecture and PCA dimension $d$. The calibration
    networks of Study~1 are $[3,3,3]$, $[5,5,5]$ and $[7,7,7]$ at $d = 10$.}
    \label{tab:occupancy_mnist}
\begin{tabular}{lrrrrrrr}
\toprule
Architecture & $d = 2$ & $d = 3$ & $d = 4$ & $d = 5$ & $d = 10$ & $d = 15$ & $d = 20$ \\
\midrule
$[3,3,3]$ & -- & -- & -- & -- & 0.004 & -- & -- \\
$[5,5,5]$ & -- & -- & -- & -- & 0.030 & -- & -- \\
$[7,7,7]$ & 0.012 & 0.028 & 0.047 & 0.082 & 0.122 & 0.125 & 0.103 \\
$[15,15,15]$ & 0.049 & 0.195 & 0.342 & 0.518 & 0.743 & 0.774 & 0.783 \\
$[25,25,25]$ & 0.124 & 0.452 & 0.734 & 0.874 & 0.954 & 0.965 & 0.979 \\
$[50,50,50]$ & 0.317 & 0.826 & 0.960 & 0.987 & 0.998 & 0.999 & 1.000 \\
\bottomrule
\end{tabular}

\end{table}

\FloatBarrier
\subsection{Bias corrections}
\label{app:bias_corrections}

The plug-in estimator is biased upwards at finite $N$. We compare it with
four standard corrections, each applied to the same contingency table:
Miller--Madow~\citep{paninskiEstimationEntropyMutual2003a}, which subtracts
$(|\Omega_{\mathcal{D}}|-1)(|\mathcal{Y}|-1)/(2N\ln 2)$;
Grassberger~\citep{grassbergerEntropyEstimatesInsufficient2003};
Chao--Shen~\citep{chaoNonparametricEstimationShannon2003} and
Chao--Wang--Jost~\citep{chaoEntropySpeciesAccumulation2013}, which are
designed for undersampled categorical data. The last three
are computed through the decomposition
$\hat H(\Omega) + \hat H(Y) - \hat H(\Omega, Y)$ with each entropy corrected
separately. All four are implemented in the released code and tested
against the \texttt{infomeasure} package~\citep{buthInfomeasureComprehensivePython2025}.
We do not report the asymptotic NSB
estimator~\citep{nemenmanCoincidencesEstimationEntropies2011}. It relies on coincidences among the samples, and their number vanishes as $\rho\to1$, so it is unstable in the regime where a correction would matter. It is computed in the released code.
\Cref{tab:bias_by_rho} averages over the
$\res{mm.n_cells}$ (dataset, architecture, PCA dimension, layer) cells of
both studies, grouped by $\rho$.

\begin{table}[h]
    \centering
    \small
    \caption{Bias corrections grouped by occupancy $\rho$ ($n$: number of
    cells). Value: corrected estimate; correction: plug-in minus corrected
    value; seed std: standard deviation over the five seeds; gap to $H(Y)$:
    $H(Y)$ minus the corrected value. All in bits, averaged over cells.}
    \label{tab:bias_by_rho}
\begin{tabular}{llrrrr}
\toprule
$\rho$ range & Estimator & Value & Correction & Seed std & Gap to $H(Y)$ \\
\midrule
$\rho < 0.05$ ($n=41$) & Plug-in & 1.439 & 0.000 & 0.148 & 1.698 \\
 & Miller--Madow & 1.325 & 0.114 & 0.131 & 1.812 \\
 & Grassberger & 1.410 & 0.029 & 0.144 & 1.727 \\
 & Chao--Shen & 1.407 & 0.032 & 0.143 & 1.730 \\
 & Chao--Wang--Jost & 1.407 & 0.032 & 0.143 & 1.730 \\
\midrule
$\rho > 0.3$ ($n=56$) & Plug-in & 2.586 & 0.000 & 0.025 & 0.146 \\
 & Miller--Madow & $-$1.153 & 3.739 & 0.092 & 3.886 \\
 & Grassberger & 2.519 & 0.067 & 0.031 & 0.214 \\
 & Chao--Shen & 2.447 & 0.139 & 0.037 & 0.285 \\
 & Chao--Wang--Jost & 2.402 & 0.185 & 0.055 & 0.331 \\
\midrule
$\rho > 0.9$ ($n=19$) & Plug-in & 2.941 & 0.000 & 0.002 & 0.004 \\
 & Miller--Madow & $-$2.462 & 5.403 & 0.038 & 5.407 \\
 & Grassberger & 2.935 & 0.006 & 0.004 & 0.010 \\
 & Chao--Shen & 2.888 & 0.054 & 0.027 & 0.058 \\
 & Chao--Wang--Jost & 2.876 & 0.065 & 0.037 & 0.069 \\
\bottomrule
\end{tabular}

\end{table}

Three regimes appear. At small occupancy ($\rho < 0.05$) every correction
is small (Miller--Madow: $\res{mm.rho<0.05.correction}$ bits on
average) compared with the seed-to-seed variability of the plug-in value
($\res{mm.rho<0.05.plug_in_std}$ bits), so the plug-in estimate is reliable. At
$\rho > 0.3$ the Miller--Madow correction grows to $\res{mm.rho>0.3.correction}$
bits on average, up to $\res{mm.max_correction}$ bits, and the corrected value
becomes negative (as low as $\res{mm.min_value}$ bits): a first-order
correction applied far outside its regime. The seed-to-seed spread of the
plug-in value moves the other way, from $\res{mm.rho<0.05.plug_in_std}$ bits
at $\rho < 0.05$ to $\res{mm.rho>0.3.plug_in_std}$ bits at $\rho > 0.3$: the
estimate is most stable where it is most biased, so its variance gives no
warning of its bias. At $\rho > 0.9$
($\res{beyond_mm.rho>0.9.n_cells}$ cells) the other corrections are small, but
for the wrong reason: their corrected values stay within
$\res{beyond_mm.rho>0.9.gap_to_H_Y.grassberger}$ (Grassberger),
$\res{beyond_mm.rho>0.9.gap_to_H_Y.chao_shen}$ (Chao--Shen) and
$\res{beyond_mm.rho>0.9.gap_to_H_Y.chao_wang_jost}$ (Chao--Wang--Jost) bits of
$H(Y)$, the same ceiling as the plug-in value. When almost every region holds
one point, the region marginal and the joint table have nearly the same
support, so the corrections of $\hat H(\Omega)$ and $\hat H(\Omega, Y)$ are
nearly equal and cancel in the difference. Miller--Madow fails visibly,
Grassberger and the coverage-based corrections fail silently. No statistical correction we tried turns a
saturated partition into an informative estimate; this is why we treat
$\rho$ as a diagnostic to report and use the functional quotient as a
structural remedy, which works where the network has redundant linear
operators to collapse (\cref{sec:capacity-results}). \Cref{tab:bias_by_rho_compact} in the main text summarises
the corrections.

\subsection{Held-out versus in-sample estimation}
\label{app:heldout}

All reported estimates use held-out points. To check how much this choice
matters, we recompute the routing estimator in-sample, on all points of
Composite ($N = \res{heldout.composite.N_insample}$) and WBC
($N = \res{heldout.wbc.N_insample}$), for the same trained networks
(\cref{tab:heldout}). At the deepest layer the plug-in estimates differ by at
most $\res{heldout.composite.max_abs_delta_plug_in.deepest}$ bits on Composite
and $\res{heldout.wbc.max_abs_delta_plug_in.deepest}$ bits on WBC (over all
layers: $\res{heldout.composite.max_abs_delta_plug_in.all_layers}$ and
$\res{heldout.wbc.max_abs_delta_plug_in.all_layers}$ bits), even though $\rho$
is lower at the larger $N$. The estimate is therefore not an
artefact of the particular points used to build the partition. For the two
saturated WBC architectures this agreement is weaker evidence: once an
estimator sits at its ceiling, its value no longer moves with $N$.

\begin{table}[h]
    \centering
    \small
    \caption{Held-out versus in-sample routing estimates at the deepest layer
    (mean over seeds, bits). Held-out / in-sample $N$: Composite
    $\res{heldout.composite.N_heldout}$ / $\res{heldout.composite.N_insample}$, WBC
    $\res{heldout.wbc.N_heldout}$ / $\res{heldout.wbc.N_insample}$. $\Delta\hat I$:
    in-sample minus held-out.}
    \label{tab:heldout}
\begin{tabular}{llrrrrr}
\toprule
Dataset & Architecture & $\rho$ held-out & $\rho$ in-sample & $\hat I$ held-out & $\hat I$ in-sample & $\Delta \hat I$ \\
\midrule
Composite & $[5,5,5]$ & 0.030 & 0.007 & 1.949 & 1.939 & $-$0.010 \\
 & $[5,5,5,5,5]$ & 0.054 & 0.013 & 2.261 & 2.245 & $-$0.015 \\
 & $[9,9,9]$ & 0.064 & 0.016 & 2.188 & 2.174 & $-$0.014 \\
 & $[9,9,9,9,9]$ & 0.147 & 0.043 & 2.550 & 2.538 & $-$0.012 \\
 & $[25,25,25]$ & 0.304 & 0.104 & 2.535 & 2.515 & $-$0.020 \\
 & $[25,25,25,25,25]$ & 0.411 & 0.163 & 2.620 & 2.609 & $-$0.011 \\
\midrule
WBC & $[5,5,5]$ & 0.184 & 0.068 & 0.767 & 0.765 & $-$0.002 \\
 & $[5,5,5,5,5]$ & 0.202 & 0.101 & 0.707 & 0.718 & 0.011 \\
 & $[9,9,9]$ & 0.360 & 0.187 & 0.740 & 0.727 & $-$0.012 \\
 & $[9,9,9,9,9]$ & 0.561 & 0.316 & 0.904 & 0.885 & $-$0.020 \\
 & $[25,25,25]$ & 0.900 & 0.741 & 0.949 & 0.919 & $-$0.031 \\
 & $[25,25,25,25,25]$ & 0.961 & 0.893 & 0.949 & 0.943 & $-$0.006 \\
\bottomrule
\end{tabular}

\end{table}

\subsection{Sensitivity of the baselines to their hyperparameters}
\label{app:baseline_sensitivity}

The calibration in \cref{sec:calibration-scatter} uses one setting per
baseline. \Cref{tab:baseline_sensitivity} repeats the correlation with the
routing estimator for every setting we computed, both for the plug-in and the
Miller--Madow-corrected binning and $k$-means estimates. On Composite and WBC
the correlation is high for every setting except the finer Miller--Madow
binnings ($16$ and $30$ bins; lowest $r = \res{sensitivity.composite+wbc.r_min}$),
and among the Miller--Madow binnings $K = 8$ is not the best-correlated bin
count ($K = 4$ is). On MNIST FC the conclusion
depends on the baseline's hyperparameter: Miller--Madow binning with $16$ and
$30$ bins per coordinate correlates negatively with the routing estimator
($r = \res{sensitivity.mnist.r.binning16}$ and $\res{sensitivity.mnist.r.binning30}$), because the correction for the many
occupied bins dominates the estimate, while the plug-in variants of the same
binnings stay positively correlated. This dependence on a free parameter is
the problem the routing estimator avoids.

\begin{table}[h]
    \centering
    \small
    \caption{Pearson $r$ between the routing estimator and each baseline
    setting, one point per network at the last hidden layer. MM:
    Miller--Madow-corrected. KSG is not a plug-in estimator and has a single
    column. $^*$The setting used in \cref{fig:calibration-scatter}.}
    \label{tab:baseline_sensitivity}
\begin{tabular}{llrrrr}
\toprule
Baseline & Setting & \multicolumn{2}{c}{Composite + WBC ($n=60$)} & \multicolumn{2}{c}{MNIST FC ($n=15$)} \\
\cmidrule(lr){3-4}\cmidrule(lr){5-6}
 &  & plug-in & MM & plug-in & MM \\
\midrule
Binning ($K$) & 2 & 0.923 & 0.946 & 0.877 & 0.875 \\
 & 4 & 0.966 & 0.973 & 0.758 & 0.728 \\
 & 8$^*$ & 0.981 & 0.945 & 0.807 & 0.690 \\
 & 16 & 0.982 & 0.852 & 0.835 & $-$0.570 \\
 & 30 & 0.975 & 0.655 & 0.824 & $-$0.888 \\
\midrule
$k$-means ($K$) & KY$^*$ & 0.970 & 0.970 & 0.720 & 0.720 \\
 & 2KY & 0.974 & 0.974 & 0.866 & 0.866 \\
 & 4KY & 0.974 & 0.974 & 0.868 & 0.868 \\
 & 16 & 0.976 & 0.976 & 0.834 & 0.834 \\
 & 64 & 0.972 & 0.971 & 0.861 & 0.861 \\
 & 256 & 0.971 & 0.964 & 0.829 & 0.829 \\
\midrule
KSG ($k$) & 3$^*$ & 0.944 &  & 0.759 &  \\
 & 5 & 0.948 &  & 0.762 &  \\
 & 10 & 0.953 &  & 0.772 &  \\
\bottomrule
\end{tabular}

\end{table}

\subsection{KSG on networks with collapsed representations}
\label{app:ksg_ties}

On five WBC networks the KSG baseline in \cref{fig:calibration-scatter}
reads zero or nearly zero while the routing estimator and the binning
baseline report $0.2$--$0.9$ bits (\cref{tab:ksg_ties}). In each of them a
large group of held-out points, $\res{ksg_ties.tied_frac_min.pct}$--$\res{ksg_ties.tied_frac_max.pct}\%$ of the $\res{ksg_ties.N}$, shares one
identical representation at the last hidden layer. Every point in such a
group passes through an earlier layer in which all units are inactive, so
everything downstream of that layer is the same constant vector. In four of
the five networks the group is almost pure in one class.

The tie breaks the KSG estimator rather than the network. For a tied point
the $k$-th nearest same-class neighbour lies at distance zero, so the
neighbourhood the estimator counts contains the whole group instead of $k$
points. This violates the assumption behind the estimator and biases it
downwards: the raw estimate is negative on four of the networks, and the
reported value is clipped at zero, as for every KSG value in the paper. The
routing estimator is unaffected, because a tied group simply contributes its
label counts to the regions it occupies.

The effect on the calibration is small. Removing the five networks leaves
the pooled KSG correlation of \cref{fig:calibration-scatter} unchanged
($\res{ksg_ties.fig2.r_ksg3.with}$ with them, $\res{ksg_ties.fig2.r_ksg3.without}$ without). In the WBC panel of
\cref{fig:layer-profile}, four of them ($[5,5,5,5,5]$, seeds $101$, $103$,
$104$, $105$) are tied from layer~$\res{ksg_ties.fig3.wbc.onset_layer_min}$ onwards and pull the mean KSG curve
down from $\res{ksg_ties.fig3.wbc.ksg3.with.max}$ to $\res{ksg_ties.fig3.wbc.ksg3.with.min}$ bits; without them, KSG stays between
$\res{ksg_ties.fig3.wbc.ksg3.without.min}$ and $\res{ksg_ties.fig3.wbc.ksg3.without.max}$ bits at every layer while $\hat I$ rises from $\res{ksg_ties.fig3.wbc.routing.without.l1}$ to
$\res{ksg_ties.fig3.wbc.routing.without.l5}$ bits. The next-largest tied share among the WBC networks is $\res{ksg_ties.untied.wbc.tied_frac_max.pct}\%$
($[5,5,5]$, seed $\res{ksg_ties.untied.wbc.seed}$, KSG $\res{ksg_ties.untied.wbc.ksg3}$ bits); on Composite and MNIST FC it is at
most $\res{ksg_ties.untied.composite.tied_frac_max.pct}\%$ and $\res{ksg_ties.untied.mnist.tied_frac_max.pct}\%$, with no visible effect on KSG.

\begin{table}[h]
    \centering
    \small
    \setlength{\tabcolsep}{4pt}
    \caption{WBC networks on which KSG collapses (last hidden layer, last
    epoch, held-out points, $N = \res{ksg_ties.N}$). Binning: Miller--Madow, $K = 8$.
    KSG raw: the estimate before clipping at zero. Tied points: size of the
    largest group of points with identical representation, and its share of
    $N$. MI in bits.}
    \label{tab:ksg_ties}
\begin{tabular}{lrrrrrrr}
\toprule
Architecture & Seed & Test acc. & $\hat I$ & Binning & KSG raw & KSG & Tied points \\
\midrule
$[5,5,5,5,5]$ & 101 & 98.2\% & 0.791 & 0.736 & $-0.469$ & 0.000 & 41 (36\%) \\
$[5,5,5,5,5]$ & 103 & 97.4\% & 0.758 & 0.795 & $0.148$ & 0.148 & 26 (23\%) \\
$[5,5,5,5,5]$ & 104 & 63.2\% & 0.392 & 0.217 & $-3.070$ & 0.000 & 89 (78\%) \\
$[5,5,5,5,5]$ & 105 & 97.4\% & 0.911 & 0.576 & $-1.743$ & 0.000 & 64 (56\%) \\
$[5,5,5]$ & 104 & 96.5\% & 0.779 & 0.720 & $-1.161$ & 0.000 & 53 (46\%) \\
\bottomrule
\end{tabular}

\end{table}

\subsection{Routing information without learned structure: label permutation}
\label{app:label_permutation}

High routing information need not reflect learned class structure. As
$\rho \to 1$ the partition approaches an injective map of the samples, and
$\hat I$ approaches $H(Y)$ for any labelling. To separate the two, we train
the $[5,5,5]$ and $[25,25,25]$ architectures on Composite and WBC with the
training labels randomly permuted, and estimate the routing information
about the \emph{true} labels on the held-out points (\cref{tab:label_permutation}).
The permuted networks learn neither the true labels nor the permuted ones:
on Composite their accuracy is near chance (test accuracy on true labels
$\res{permutation.composite.[5, 5, 5].permuted.test_accuracy.pct}\%$ and
$\res{permutation.composite.[25, 25, 25].permuted.test_accuracy.pct}\%$, accuracy
on their own permuted training labels
$\res{permutation.composite.[5, 5, 5].permuted.train_label_accuracy.pct}\%$ and
$\res{permutation.composite.[25, 25, 25].permuted.train_label_accuracy.pct}\%$), and on
WBC they predict close to the majority class.

For the narrow Composite network the routing information drops from
$\res{permutation.composite.[5, 5, 5].clean.raw}$ to
$\res{permutation.composite.[5, 5, 5].permuted.raw}$ bits, and the functional
quotient discounts it further
($\res{permutation.composite.[5, 5, 5].clean.quotient}$ to
$\res{permutation.composite.[5, 5, 5].permuted.quotient}$ bits). For the wide
network it barely moves: $\res{permutation.composite.[25, 25, 25].clean.raw}$
to $\res{permutation.composite.[25, 25, 25].permuted.raw}$ bits raw and
$\res{permutation.composite.[25, 25, 25].clean.quotient}$ to
$\res{permutation.composite.[25, 25, 25].permuted.quotient}$ bits after the
quotient, close to $H(Y) = \res{hy.composite.heldout}$ bits, although the
network predicts the true label of only
$\res{permutation.composite.[25, 25, 25].permuted.test_accuracy.pct}\%$ of the test
points. A fine partition of Composite's spatially coherent classes correlates
with the true class whether or not the network has learned it. On WBC the
same holds for $[25,25,25]$, whose permuted network sits at $\rho > 0.9$.
High routing information is therefore not by itself evidence of learned
structure, and the functional quotient, while a valid lower bound by the
data-processing inequality, removes this effect only at moderate capacity.

\begin{table}[h]
    \centering
    \small
    \setlength{\tabcolsep}{4pt}
    \caption{Clean versus label-permuted training (deepest layer, held-out
    points, true labels, mean over seeds). Test acc.: accuracy on the true
    test labels; train-label acc.: accuracy on the (permuted) labels the
    network was trained on. $\hat I_{\mathrm{func}}$ and
    $\rho_{\mathrm{func}}$ at $\varepsilon = 0.3$. MI in bits; $H(Y) =
    \res{hy.composite.heldout}$ (Composite) and $\res{hy.wbc.heldout}$ (WBC).}
    \label{tab:label_permutation}
\begin{tabular}{lllrrrrrr}
\toprule
Dataset & Architecture & Labels & Test acc. & Train-label acc. & $\rho$ & $\hat I$ & $\hat I_{\mathrm{func}}$ & $\rho_{\mathrm{func}}$ \\
\midrule
Composite & $[5,5,5]$ & clean & 97.9\% & -- & 0.030 & 1.949 & 1.829 & 0.765 \\
 & $[5,5,5]$ & permuted & 19.6\% & 20.8\% & 0.022 & 1.411 & 1.096 & 0.656 \\
 & $[25,25,25]$ & clean & 99.5\% & -- & 0.304 & 2.535 & 2.362 & 0.377 \\
 & $[25,25,25]$ & permuted & 15.9\% & 22.5\% & 0.245 & 2.433 & 2.329 & 0.645 \\
\midrule
WBC & $[5,5,5]$ & clean & 97.4\% & -- & 0.184 & 0.767 & 0.700 & 0.680 \\
 & $[5,5,5]$ & permuted & 62.5\% & 62.8\% & 0.211 & 0.469 & 0.295 & 0.709 \\
 & $[25,25,25]$ & clean & 97.2\% & -- & 0.900 & 0.949 & 0.885 & 0.745 \\
 & $[25,25,25]$ & permuted & 63.3\% & 65.6\% & 0.933 & 0.946 & 0.932 & 0.910 \\
\bottomrule
\end{tabular}

\end{table}

\subsection{Sensitivity of the functional quotient to the visiting order}
\label{app:ordering}

The greedy clustering of \cref{alg:func_cluster} depends on the order in
which regions are visited (\cref{app:func_algorithm}). For every clean
network and layer we rerun it with $\res{ordering.n_random_orders}$ random
orders in addition to the first-encounter order and record the spread
(maximum minus minimum) of the quotient size and of the Miller--Madow
quotient MI (\cref{tab:ordering}). At $\varepsilon = 0.3$ the median spread is
$\res{ordering.eps0.3.all_layers.spread_median}$ classes; the largest is
$\res{ordering.eps0.3.all_layers.spread_max}$ classes, or
$\res{ordering.eps0.3.all_layers.relative_spread_max.pct}\%$ of
$|\Omega_{\mathcal{D}}|$, and the quotient MI varies by at most
$\res{ordering.eps0.3.all_layers.miller_madow_func_spread_max}$ bits
($\res{ordering.eps0.3.deepest.miller_madow_func_spread_max}$ bits at the
deepest layer). The spread grows with $\varepsilon$: over all four values,
$\varepsilon \in \{0.1, 0.3, 0.5, 1\}$, it reaches $\res{ordering.all_eps.miller_madow_func_spread_max}$ bits. The order
is thus a small but not negligible source of variation at large
$\varepsilon$, which is one more reason to stay in the moderate range.

\begin{table}[h]
    \centering
    \small
    \caption{Spread of the functional quotient over visiting orders, per
    $\varepsilon$. Cells: (network, layer) pairs. Median, 95th percentile and
    maximum of the spread of the quotient size; max$/R$: largest spread
    relative to $|\Omega_{\mathcal{D}}|$; last column: largest spread of the
    Miller--Madow quotient MI (bits).}
    \label{tab:ordering}
\begin{tabular}{lrrrrrrr}
\toprule
Layers & $\varepsilon$ & Cells & Median & 95th pct. & Max & Max / $R$ & Max $\Delta \hat I^{\mathrm{MM}}_{\mathrm{func}}$ \\
\midrule
all layers & 0.1 & 690 & 0 & 9.0 & 17 & 6.7\% & 0.053 \\
 & 0.3 & 690 & 2 & 38.5 & 84 & 9.4\% & 0.148 \\
 & 0.5 & 690 & 4 & 43.0 & 83 & 18.8\% & 0.348 \\
 & 1 & 690 & 1 & 19.0 & 47 & 14.3\% & 0.474 \\
\midrule
deepest layer & 0.1 & 210 & 0 & 11.6 & 17 & 6.2\% & 0.053 \\
 & 0.3 & 210 & 11 & 45.5 & 60 & 6.5\% & 0.100 \\
 & 0.5 & 210 & 14 & 53.1 & 83 & 8.3\% & 0.348 \\
 & 1 & 210 & 5 & 28.0 & 47 & 10.5\% & 0.474 \\
\bottomrule
\end{tabular}

\end{table}

\newpage
\section{Implementation, Assets and Compute}
\label{app:implementation}

\paragraph{Code.} All code, the stored results and instructions to
reproduce every figure and table are released under the MIT licence at
\url{https://github.com/Johanmkr/Intrinsic-IT-analysis-of-ReLU-nets}, with a
locked environment (\texttt{uv.lock}) and a Dockerfile. Networks are trained
with PyTorch~2.11.0 (CPU build) \citep{paszkePyTorchImperativeStyle2019}; the
Composite generators (\texttt{make\_moons}, \texttt{make\_circles}), PCA,
min-max scaling and $k$-means come from scikit-learn~1.8.0
\citep{pedregosaScikitlearnMachineLearning2011}.

\paragraph{Baselines.} Binning quantises every unit of $T$ into $K$ uniform
bins on $[-c, c]$, with $c$ the largest absolute activation in the layer, and
uses the joint bin index as a discrete variable. $k$-means uses scikit-learn
with $10$ initialisations, seeded by the network seed. KSG is our own
implementation of the mixed continuous--discrete estimator
\citep{rossMutualInformation2014} with $k = 3$; negative values are clipped at
zero (\cref{app:ksg_ties}). Binning and $k$-means are scored with the plug-in
and the Miller--Madow estimators (\cref{app:baseline_sensitivity}).

\paragraph{Licences of existing assets.} MNIST \citep{lecun1999mnist}:
CC BY-SA 3.0. WBC \citep{williamwolbergBreastCancerWisconsin1993}: CC BY 4.0
(UCI Machine Learning Repository). PyTorch, scikit-learn, NumPy and SciPy:
BSD-style licences. The \texttt{infomeasure} package~0.6.2
\citep{buthInfomeasureComprehensivePython2025} (AGPL-3.0) is used only in
tests of our bias-correction implementations and is not part of the default
installation.

\paragraph{Compute.} Everything runs on CPU. On a 22-core x86-64
workstation with $20$ parallel single-threaded workers, the full pipeline
takes about $2$\,h of wall time, about $42$ CPU-hours: training the $230$
networks of both studies and the label-permutation control takes $71$ min
(about $24$ CPU-hours, mostly the MNIST networks at about $3.5$ min each),
the routing estimates $39$ min, the baselines $8$ min and the diagnostics
$6$ min. Memory use is small. The full research project required additional
CPU compute for preliminary experiments and earlier versions of the pipeline
that are not reported.

\newpage
\section{Training Metrics}
    \label{app:train_metrics}

\begin{figure}[h!]
      \centering
      \includegraphics[width=\textwidth]{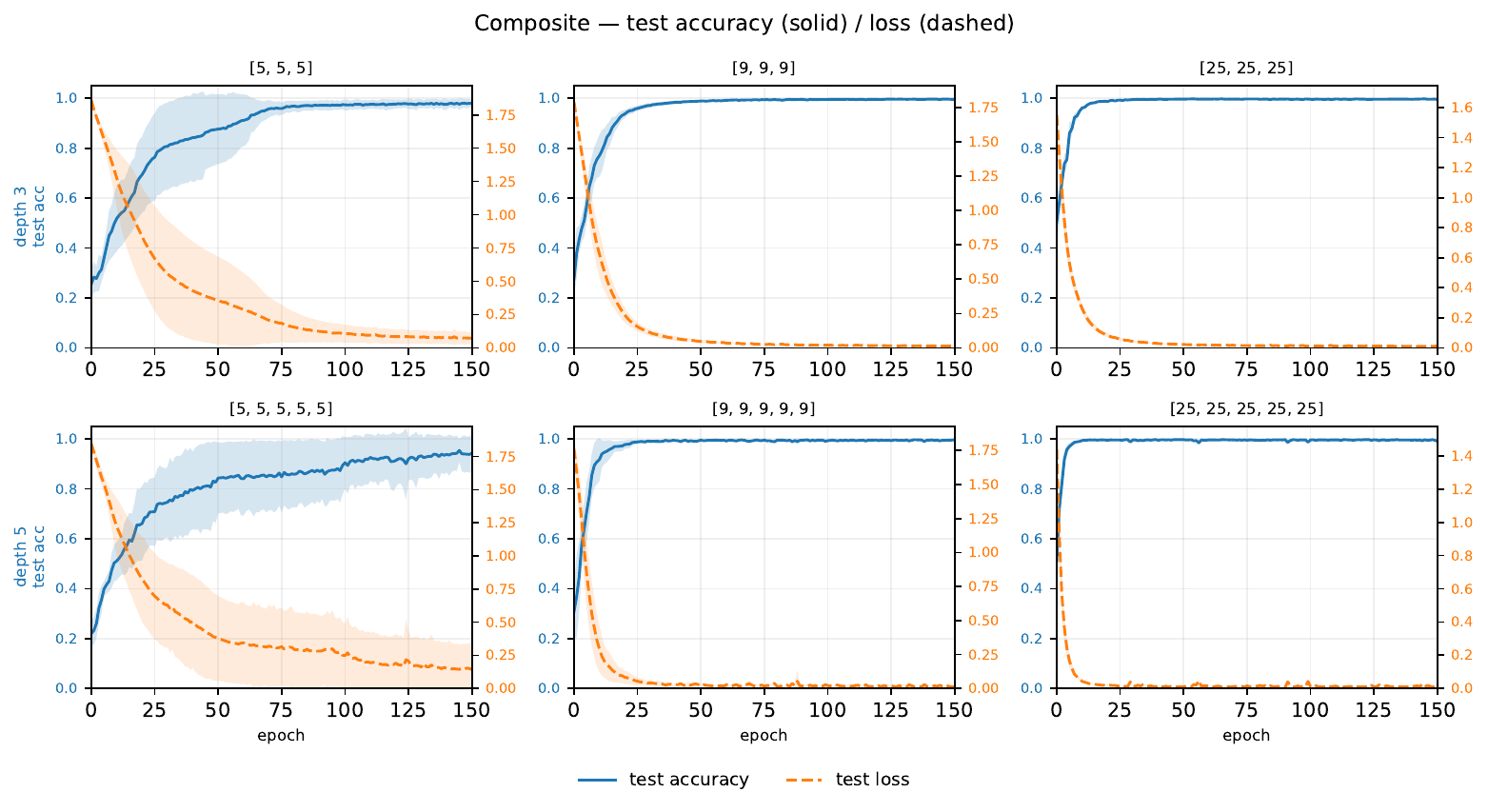}
      \caption{Composite: test accuracy (solid) and test loss (dashed) of
the six Study-1 architectures, mean $\pm 1\sigma$ over five seeds.}
      \label{fig:composite_metrics}
    \end{figure}

\begin{figure}[h!]
      \centering
      \includegraphics[width=\textwidth]{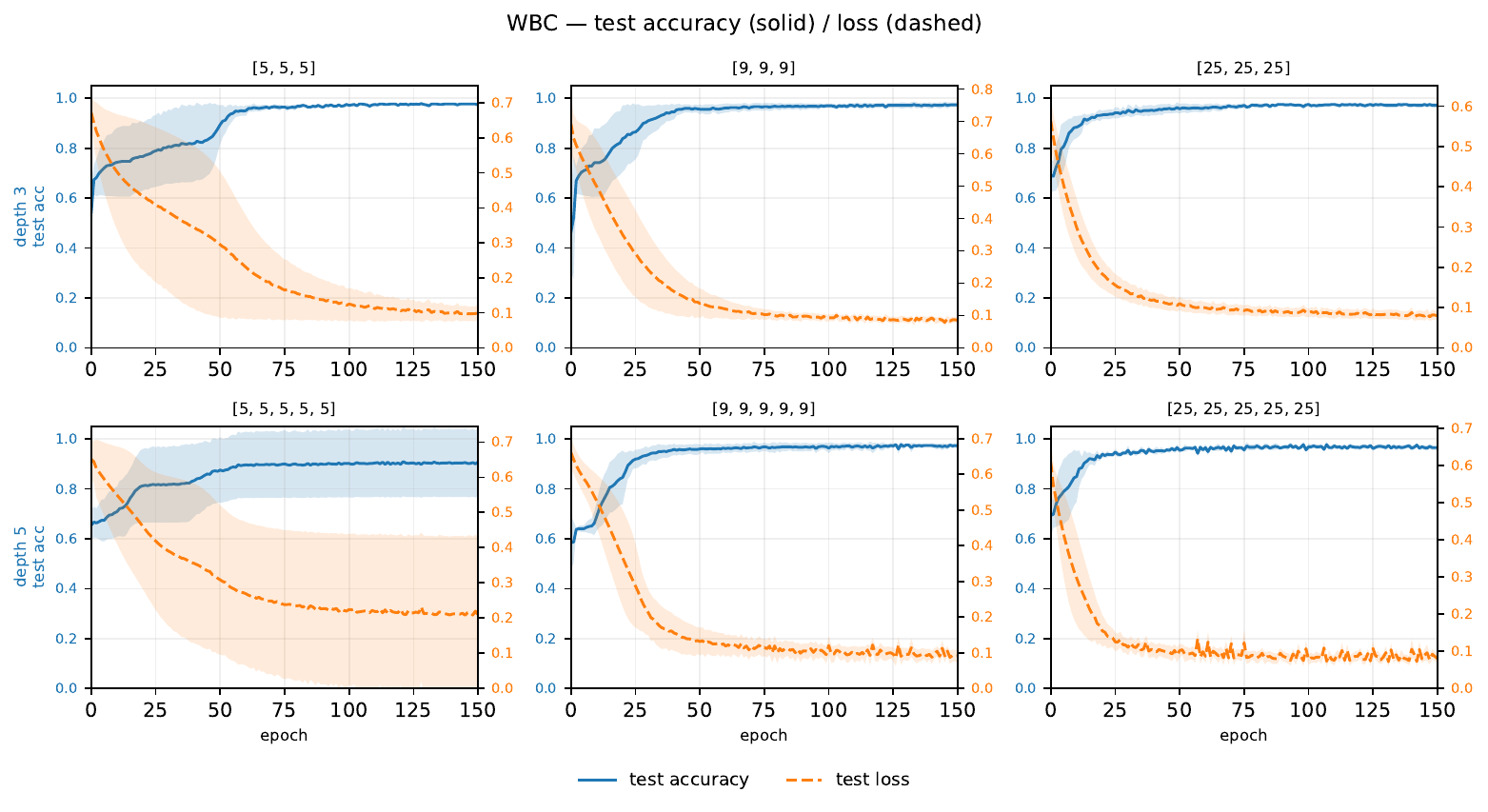}
      \caption{WBC: test accuracy (solid) and test loss (dashed) of the
six Study-1 architectures, mean $\pm 1\sigma$ over five seeds.}
      \label{fig:wbc_metrics}
    \end{figure}

\begin{figure}[t]
      \centering
      \includegraphics[width=\textwidth]{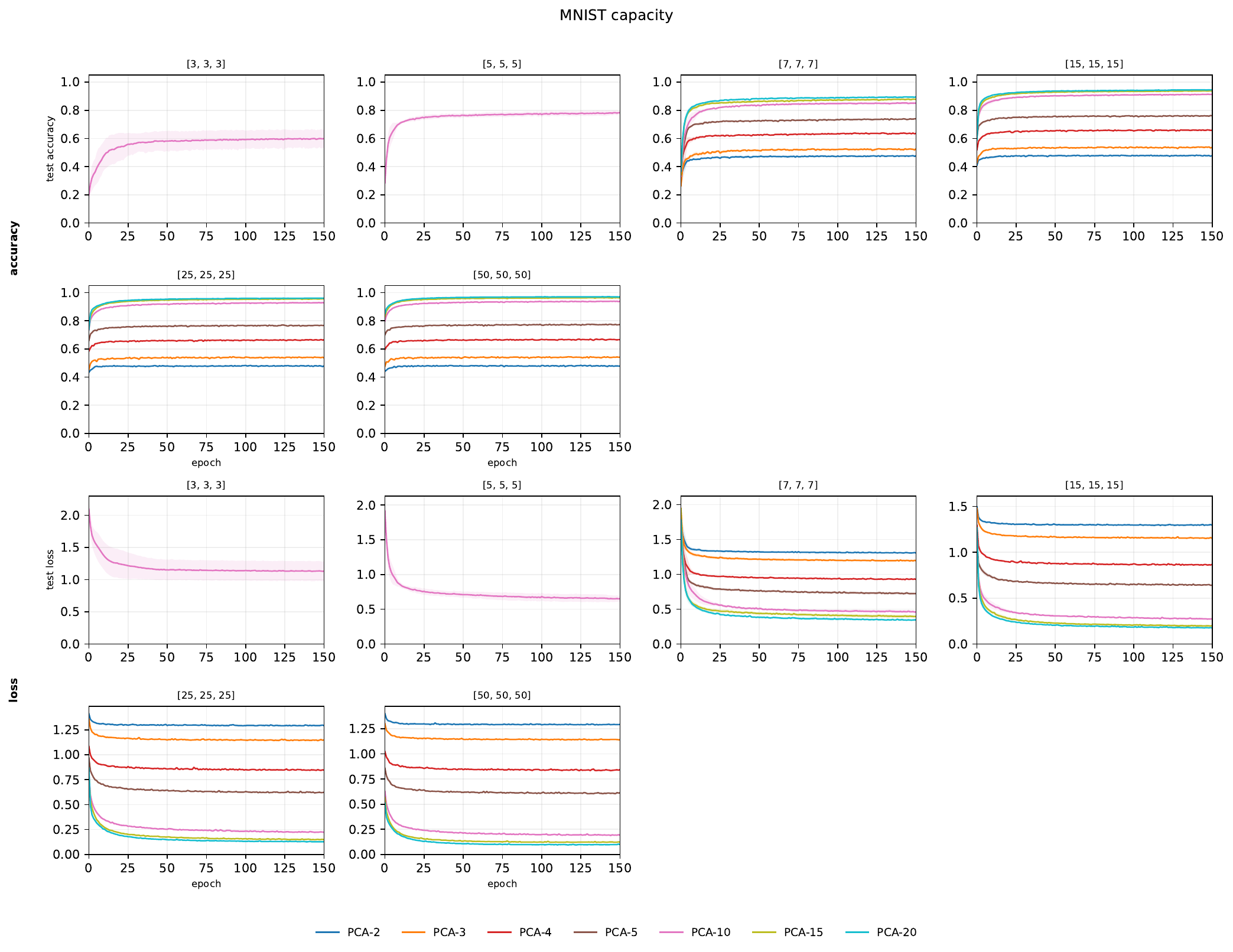}
      \caption{MNIST: test accuracy (top two rows) and test loss (bottom
two rows), one colour per PCA dimension $d$. $[3,3,3]$ and $[5,5,5]$
are the Study-1 networks at $d = 10$; $[7,7,7]$ to $[50,50,50]$ are the
Study-2 networks at every $d$ ($[7,7,7]$ at $d = 10$ also enters
Study~1). Mean $\pm 1\sigma$ over five seeds.}
      \label{fig:mnist_metrics}
    \end{figure}

    Test accuracy and test cross-entropy loss over training, on the
held-out points, for every network in the paper: the six Composite
and six WBC architectures of Study~1
(\cref{fig:composite_metrics,fig:wbc_metrics}) and all MNIST networks
of both studies (\cref{fig:mnist_metrics}). Curves are means over the
five seeds, shaded bands $\pm 1\sigma$.

\paragraph{Composite}

    \paragraph{WBC}
One network does not converge: seed $104$ of $[5,5,5,5,5]$ ends at
$63\%$ test accuracy, close to the majority-class rate of $62.7\%$,
which produces the wide band in that panel.

    \newpage
    \paragraph{MNIST}


\end{document}